\documentclass[10pt,twocolumn,letterpaper]{article}
\usepackage[pagenumbers]{new}
\usepackage{graphicx}
\usepackage{amsmath}
\usepackage{amssymb}
\usepackage{booktabs}
\usepackage{bbding}
\usepackage{pifont}
\usepackage{wasysym}
\usepackage{arydshln}
\usepackage{multirow}
\usepackage{float}
\usepackage{csquotes}
\usepackage{rotating}
\usepackage{times}
\usepackage{epsfig}
\usepackage{caption}
\usepackage{subcaption}
\usepackage{setspace}
\usepackage{enumitem}
\usepackage[pagebackref,breaklinks,colorlinks]{hyperref}

\makeatletter
\def\thanks#1{\protected@xdef\@thanks{\@thanks
		\protect\footnotetext{#1}}}
\makeatother
\begin{document}
\title{TriStereoNet: A Trinocular Framework for Multi-baseline Disparity Estimation}
\author{
	Faranak Shamsafar \hspace{4cm}Andreas Zell \\ 
	{\tt\small \hspace{1.7cm} f.shmsfr@gmail.com} \hspace{2.5cm} {\tt\small andreas.zell@uni-tuebingen.de}\\
	\thanks{\hspace{-0.5cm}This work is funded by the Federal Ministry of Education and Research (BMBF) and the Baden-W{\"u}rttemberg Ministry of Science as part of the Excellence Strategy of the German Federal and State Governments.}
}
\maketitle
\begin{abstract}
While various deep learning-based approaches have been developed for stereo vision, binocular setups with fixed baselines present limited input data and face occlusion issues when covering a large depth range. We address this problem by using both narrow and wide baselines. Also, with increased evidence coming from the wide baseline, the recovered depth range can be extended. This scheme is beneficial for autonomous urban and highway driving applications. Thus, we present a model for processing the data from a trinocular setup with a narrow and a wide stereo pair. In this design, two pairs of binocular data with a common reference image are treated with one network in an end-to-end manner. We explore different methods and levels of fusion, and propose a Guided Addition module for a mid-level fusion of the two baseline data. In addition, a method is presented to iteratively optimize the network with sequential self-supervised and supervised learning on real and synthetic datasets. With this method, we can train the model on real-world data without needing the ground-truth disparity maps. Quantitative and qualitative results demonstrate that the multi-baseline network surpasses the model with a similar architecture but trained with individual baseline data. In particular, the trinocular setup outperforms the narrow and wide baselines by 13\% and  24\% in D1 error, respectively. Code and dataset: \url{https://github.com/cogsys-tuebingen/tristereonet}. 
\end{abstract}
\section{Introduction}
While there are several methods for estimating depth, stereo matching is the most adaptable to various use-cases. Over the past three decades, the passive recovery of depth maps using stereo has attracted interest in the computer vision field, and numerous real-world applications, like autonomous driving and robot navigation, can benefit from estimating depth directly from images. Other depth estimation technologies for autonomous vehicles, such as Laser Imaging Detection and Ranging (LiDAR) sensors, actively measure depth based on the travel time of a light beam emitted by the device. However, in addition to being expensive to set up, LiDAR produces a sparse depth map that is also susceptible to weather conditions. Estimating depth via stereo is done by calculating the \emph{disparities} (displacements) between the matching points in the rectified images, which makes it simple to estimate the depth via triangulation. This is usually achieved with two images (\ie a binocular setup) and different methods can be used to estimate the disparity between them.

\begin{figure}[t]
	\begin{center}
		\includegraphics[width=0.6\linewidth]{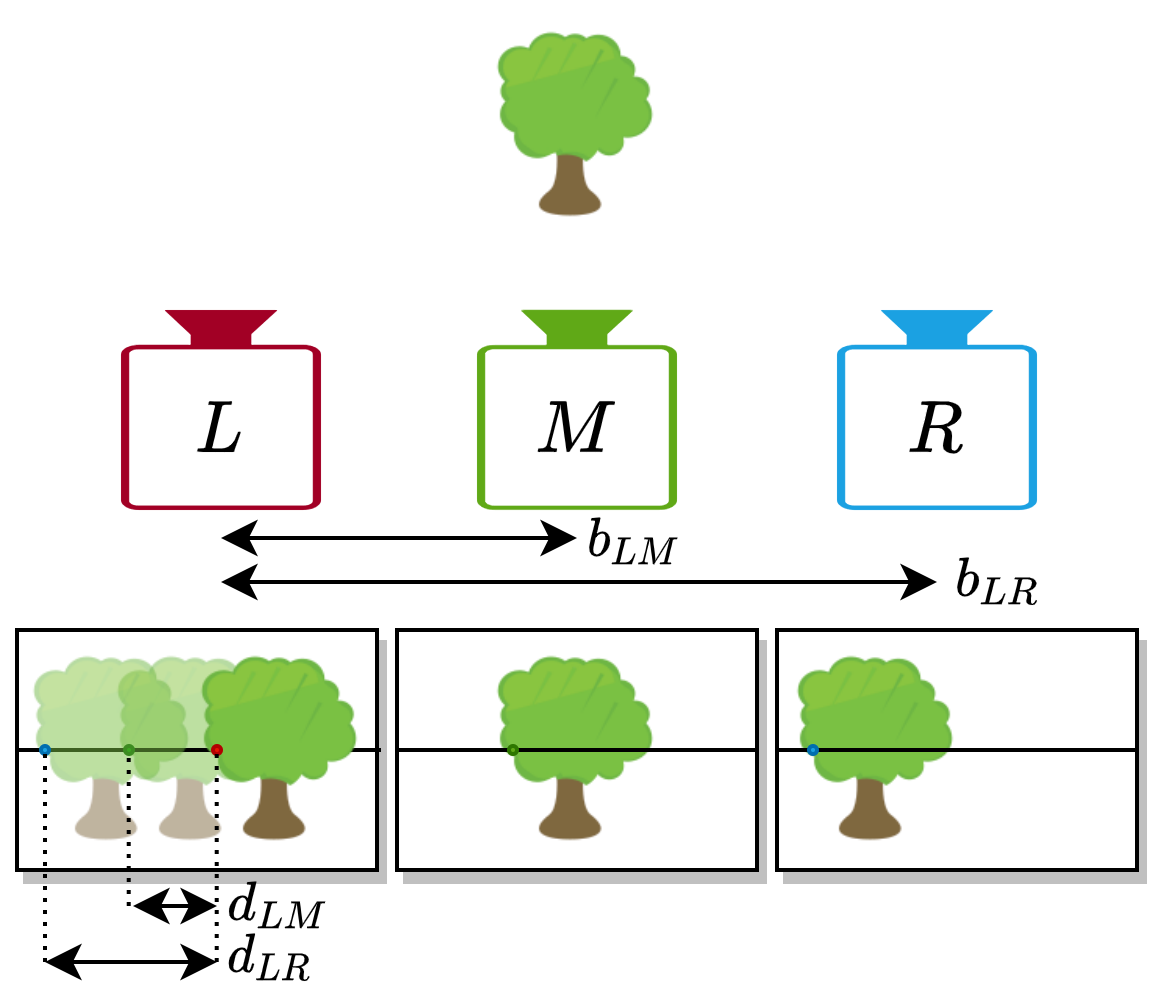}
	\end{center}
	\vspace{-0.6cm}
	\caption{Multi-baseline stereo with three cameras (\textbf{L}eft, \textbf{M}iddle, \textbf{R}ight). Note that disparity (displacement of matching points) is directly dependent on the baseline. The left image is the reference.}
	\label{fig:baseline}
\end{figure}
Before deep learning, various hand-engineered features, like Sum of Absolute Difference or Census Transform \cite{zabih1994non} were used for finding matching points and for computing a cost volume. This was followed by a regularization module, such as the well-known Semi Global Matching (SGM) method \cite{hirschmuller2005accurate}. Deep learning-based approaches, on the other hand, attempt to adapt a network for either some steps of stereo vision \cite{vzbontar2016stereo,seki2017sgm} or the entire pipeline as an end-to-end technique \cite{kendall2017end,chang2018pyramid,guo2019group,zhang2019ga}. The latter group has significantly improved the accuracy of stereo vision. 
Nevertheless, the applicability of these approaches in real-world scenarios remains a concern due to their bias toward the content of the training images, which includes varying object distances from the camera.
\begin{figure*}[t]
	\begin{center}
		\includegraphics[width=1\linewidth,page=2]{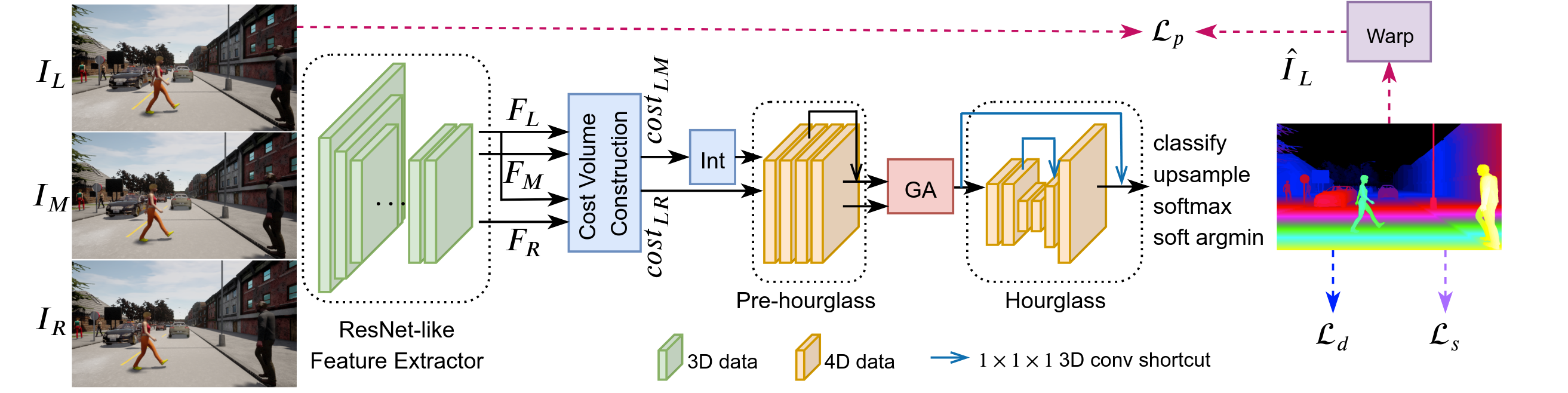}
	\end{center}
	\vspace{-0.5cm}
	\caption{Overview. After extracting the feature of the 3-tuple rectified input images ($(I_L,I_M,I_R)$ $\rightarrow$ $(F_L,F_M,F_R)$), cost volumes are computed for the narrow ($(F_L,F_M)$ $\rightarrow$ $cost_{LM}$) and wide ($(F_L,F_R)$ $\rightarrow$ $cost_{LR}$) baselines. \enquote{Int} applies the interpolation to stretch  $cost_{LM}$ across the disparity dimension. \enquote{GA} or Guided Addition fuses the two streams of data. While $\mathcal{L}_{d}$ and $\mathcal{L}_{s}$ losses are used in supervised learning, $\mathcal{L}_{p}$ and $\mathcal{L}_{s}$ are exploited for self-supervision.}
	\label{fig:generalframework}
\end{figure*}

Using multiple images to calculate depth is a possible solution as it provides more visual clues about the scene. Additionally, in some applications like autonomous driving, having more than one fixed baseline is critical. The reason is that while close-range distances require obtaining depth information via a narrow baseline, such as in low/moderate-speed urban driving settings with nearby vehicles or pedestrians, a wide baseline is equally essential when driving at high speeds on a highway. Note that depth resolution decreases quadratically with depth in a fixed baseline \cite{gallup2008variable}. More specifically, each stereo camera (with a single baseline) has a blind range or occlusion, \ie the region that cannot be seen by both cameras at the same time. The blind range of a stereo camera with a narrow baseline is smaller than that of a stereo camera with a wider baseline. A narrow stereo camera, however, is not able to accurately infer depth at a distance. By contrast, using a wider baseline leads to more accurate predictions for distant regions, but increases false matches due to the large disparity search range and severe occlusion problem. Therefore, depending on the scene and object distances, both narrow and wide baselines are needed.  

To address this issue, we propose a three-view stereo setup (Fig. \ref{fig:baseline}) and a deep network (Fig. \ref{fig:generalframework}) for estimating the disparity by three inputs. Our work considers two baselines of this trinocular setup with a shared reference image to obtain accurate disparity maps. To the best of our knowledge, the proposed model is the first that processes the multi-baseline trinocular setting in an end-to-end deep learning-based manner. Our motivation for using three cameras for stereo is to leverage both the narrow and the wide baselines for accurate depth estimation. Furthermore, in such a setup, the visual data can be recovered for objects in the blind range of the wider baseline by using the narrow baseline. This formulation is particularly necessary for autonomous navigation and driving scenarios where diversified and unpredictable near- and far-range objects appear. The main contributions of this work are summarized as follows: \emph{1)} We design a horizontally-aligned multi-baseline trinocular stereo setup for more accurate depth estimation. Accordingly, an end-to-end deep learning-based model is proposed for processing the 3-tuple input. \emph{2)} We propose a new layer for merging the disparity-related data for a mid-level fusion. We also investigate other levels and methods of fusion. \emph{3)} An iterative sequential self-supervised and supervised learning is proposed to make the design applicable to new real-world scenarios where disparity annotations are unavailable. \emph{4)} We build a synthetic dataset for the trinocular design together with the ground-truth information.
\section{Related Work} 
\label{RelatedWork}
Classical algorithms for stereo vision can mainly be divided into three modules: matching cost computation, cost regularization, and disparity optimization. For cost regularization, local approaches calculate disparities based on the neighboring pixels \cite{chen2001fast,muhlmann2002calculating} with the drawback of the sensitivity to occlusion and uniform texture. Alternatively, the global methods consider disparity changes globally to increase accuracy, but at the cost of higher computational complexity. The semi-global approaches examines both locally for predicting better disparities for small regions, as well as globally for estimating based on the overall content of the images \cite{hirschmuller2005accurate}. After the introduction of semi-global solution, classical work mainly focused on improving its accuracy and speed \cite{gehrig2007improving,michael2013real}.

With the rise of deep learning, stereo matching continued to be reformed by these modern techniques. There are two categories of deep models that follow the general paradigm for stereo reconstruction: the methods that formulate one or some of the steps with a deep learning framework \cite{vzbontar2016stereo,batsos2018cbmv,seki2017sgm}, and the approaches that transfer the full process in an end-to-end scheme \cite{mayer2016large, kendall2017end,chang2018pyramid,guo2019group,zhang2019ga,shamsafar2021mobilestereonet}. Based on recent advances, our model is end-to-end, but processes a 3-tuple sample.

The concept of multi-baseline stereo was first explored back in 1993 \cite{okutomi1993multiple} to benefit from narrow and wide baselines. In \cite{okutomi1993multiple}, the fusion was applied after computing the Sum of Squared Distances (SSD) of images. \cite{kallwies2018effective,kallwies2020triple} studied a three-view setup by adding a camera on top of the left one. In \cite{milella20143d}, the authors validate the performance of a multi-baseline system for autonomous navigation in agricultural and off-road environments. The system integrates the point clouds obtained from the wide and narrow baselines. An FPGA-based multi-baseline stereo system with four cameras was developed in \cite{honegger2017embedded} via Census Transform and without regularization like SGM. The authors showed that their setup is superior to binoculars \emph{with} SGM in recovering fine structures. In a different work \cite{wang2020stereo}, multi-baseline is used to accelerate the SGM-based disparity estimation. In contrast to these studies, which are all based on classical stereo methods, our multi-baseline design combines information from the wide and narrow baselines in a full deep learning-based framework.
\section{Methodology}
\label{ProposedMethod}
\subsection{Trinocular Stereo Setup}
The schematic of our multi-baseline trinocular setup with horizontally-aligned cameras is illustrated in Fig.\ref{fig:baseline}. Due to the fact that identical cameras are positioned in parallel with known displacements, not only can we perform stereo matching between different pairs of cameras, but we can also fuse the data for accurate and robust prediction. In our formulation, we consider two left-middle ($b_{LM}$/$b_{narrow}$) and left-right ($b_{LR}$/$b_{wide}$) baselines with the left image as reference. Note that it is possible to consider other stereo pairs or reference images. We fuse the information from the driver's perspective (for right-hand traffic) to evaluate our method in driving scenarios.

This setup is an array of axis-aligned cameras, where the matching pixels are located on the same horizontal line, allowing the fusion to take place, \ie both the LM and LR pairs contribute to disparity estimation. Theoretically, the disparities between the matching points in LM and LR pairs depend on the baselines. That is, via triangulation for a fixed object and given $b_{LR}=r\cdot b_{LM}$, we get $d_{LR}=r\cdot d_{LM}$, with $d$ as the notation for disparity and $r=2$ as the ratio between the baselines. As a result, by fusing the LM and LR pairs, more constraints are introduced since $d_{LR}=r\cdot d_{LM}$ must be satisfied. Hence, this setup benefits from both the wide (for accurate depth estimation of distant objects) and narrow baseline (for depth estimation of close objects that lie in the blind range of the wide baseline).
\subsection{Network Structure} 
\noindent\textbf{Backbone.} Our TriStereoNet architecture is depicted in Fig. \ref{fig:generalframework}, which is based on GwcNet \cite{guo2019group}. GwcNet was originally proposed for standard binocular stereo and is divided into four main modules: feature extraction, cost volume construction (which creates 4D data as the output), pre-hourglass, and hourglasses (encoder-decoder). Since the hourglass modules operate on 4D data, they are built with heavy 3D convolutions. Thus, we use only a single hourglass for efficiency.

In our modification for three inputs, the ResNet-like feature extraction \cite{chang2018pyramid, guo2019group} is shared for three rectified images ($I_L,I_M,I_R$). Given the images of size $3\times H\times W$, the feature data for each image is $320\times H/4\times W/4$. After computing the three feature data ($F_L,F_M,F_R$) for the 3-tuple sample, two cost volumes are computed for the LM and LR pairs ($cost_{LM}$ and $cost_{LR}$) by group-wise correlation \cite{guo2019group}. These 4D cost data are of size $F\times D\times H/4\times W/4$, where $F=40$ is the number of groups in group-wise correlation and $D=d_{max}/4$ is the maximum disparity range (assumed to be $d_{max}=192$ for the input image size).

\begin{figure}[t]
	\begin{center}
		\includegraphics[width=0.8\linewidth]{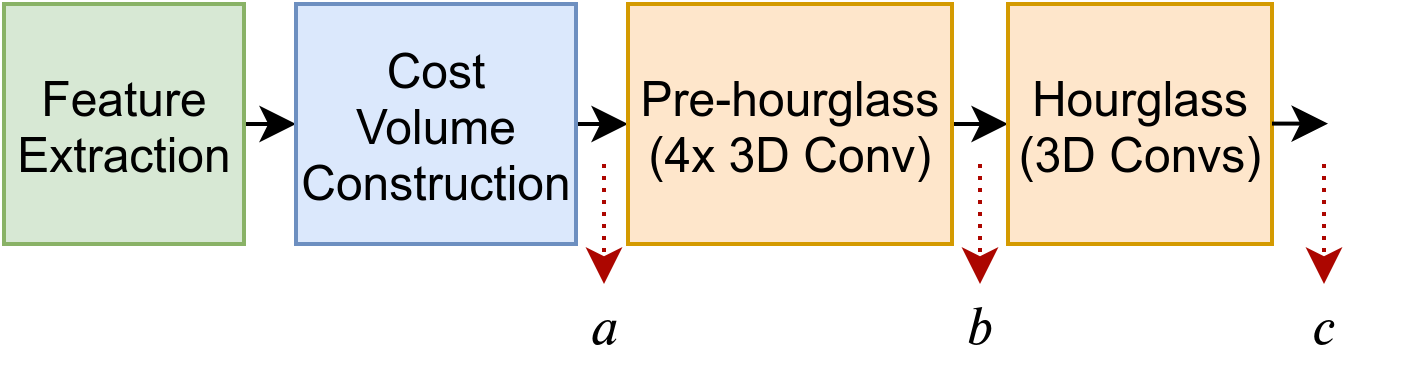}
	\end{center}
	\vspace{-0.7cm}
	\caption{Different levels for merging the narrow- and wide-baseline data: after cost volume ($a$), after pre-hourglass ($b$), and after hourglass ($c$).}
	\label{fig:levels}
\end{figure}

\begin{figure}[t]
	\begin{center}
		\includegraphics[width=0.45\linewidth]{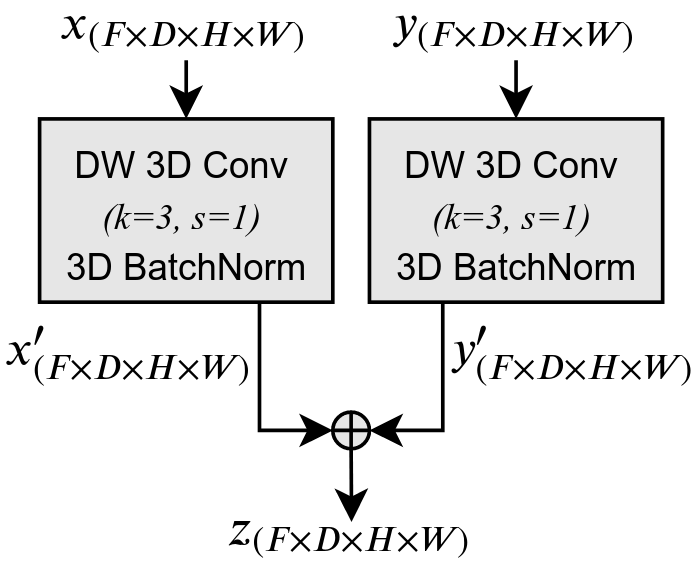}
	\end{center}
	\vspace{-0.7cm}
	\caption{The proposed Guided Addition (GA) layer to merge two 4D data. $F$, $D$, $H$-$W$ denote the feature, disparity, and spatial size, respectively.}
	\label{fig:ga}
\end{figure}

\noindent\textbf{Aligning the Cost Volumes.} One notable point is that the baselines of the LM and LR pairs are different. Thus, the disparity range of the two cost volumes must be aligned for fusion. Namely, we need to consider $cost_{LM}(f,d/r,i,j)$ to align the disparity values with $cost_{LR}(f,d,i,j)$. Here, $i,j$ stand for spatial dimension. To compute $cost_{LM}(f,d/r,i,j)$ at non-integer values, we should interpolate the values across the disparity dimension. To this end, we utilize natural cubic splines with a function as eq. \ref{eq:spline} for each sub-interval $[x_{j-1},x_{j}]$ ($j=2,3,...n$). For more details on how the related coefficients can be computed, we refer the reader to \cite{mckinley1998cubic}.
\begin{equation}
	{\displaystyle {S}_{j}\left(x\right)=a_{j}+b_{j}\left(x-x_{j}\right)+c_{j}{\left(x-x_{j}\right)}^{2}+d_{j}{\left(x-x_{j}\right)}^{3}}
	\label{eq:spline}
\end{equation}
\noindent\textbf{Fusion of Narrow- \& Wide-Baseline Data.} At this stage, the main questions are \emph{where} and \emph{how} to apply the fusion of the two baseline data. Note that an important assumption for making this fusion feasible is that all three images must be rectified, and this can be obtained by our trinocular setup. There are mainly three levels of fusion in the network (Fig. \ref{fig:levels}): \emph{1)} after cost volume computation, \emph{2)} after pre-hourglass, and \emph{3)} after hourglass. We find that, after cost volume computation, if data are processed with a few more convolutions, \ie with pre-hourglass module, their aggregation obtains higher accuracy. Fusion after hourglass also outperforms direct fusion of the cost volumes; however, the complexity increases as the network operates two 4D data (instead of one) in the hourglass with all heavy 3D convolutions. We propose a \emph{Guided Addition} (GA) module (Fig. \ref{fig:ga}) for merging two streams of the 4D disparity data \emph{after pre-hourglass}. This module applies depth-wise 3D convolution across the \emph{feature} (and not disparity) dimension to each 4D data. After 3D batch normalization, the layer merges the data by addition. This simple yet effective sub-network shows superior performance to methods like direct addition. Also, it retains the data size with kernel size 3, stride 1, and the same number of channels.

We investigated other manners and other levels of fusion. Namely, for cost volume fusion, we applied addition, average, concatenation, maximization and top feature selection (\ie getting the largest elements from the two data across the feature dimension). Also, we examined average and top feature selection for pre-hourglass and hourglass fusion. However, for accuracy and efficiency, we adopted Guided Addition after pre-hourglass (${pre\_hg_{ga}}$, \emph{c.f.} Tab. \ref{tab:comb}). We also observed that fusion by addition and average in cost level, and top feature selection in pre-hourglass and hourglass levels do not outperform the single LR baseline. This indicates the necessity of the appropriate \emph{level} and \emph{method} of fusion.

It should be noted that although our network processes in-line images, it can easily be adapted for a vertical baseline. For instance, given an L-shape trinocular setup (Top, Left, Right), with the left image as reference, we only need to rotate the top and left image features (which correspond to the vertical baseline) by 90$^{\circ}$ before cost volume, and then rotate back the resulted cost for fusion with the horizontal baseline. As the main focus of our paper is designing a deep end-to-end model for multi-baseline processing, it can be generalized to any other camera configuration as long as the images are rectified in a specific direction.
\vspace{-0.1cm}
\subsection{Learning Mechanism}
\noindent\textbf{Supervised Learning.} By generating a trinocular synthetic dataset with ground-truth information, we can apply supervised learning via a loss function between the estimated and the ground-truth disparity maps. To this end, we employ Huber loss as eq. \ref{eq:huber}, with $d$ and $\hat d$ as the ground-truth and estimated disparity maps, and $n$ as the number of image pixels. $\delta$ is the threshold for the scaled L1 and L2 loss. We show in Sec. \ref{ExperimentalResults} that the value of $\delta$ affects the performance. Previous works mostly use smooth L1 loss for disparity loss \cite{chang2018pyramid,guo2019group,zhang2019ga,duggal2019deeppruner,shen2021cfnet}.
\begin{equation}
	\mathcal{L}_d(d,\hat{d}) =\frac{1}{n}\sum_{i}
	\begin{cases} 
		0.5(d_i-\hat{d_i})^2, \text{\hspace{2.8em}if\hspace{0.2em}} |d_i-\hat{d_i}|<\delta\\
		\delta\times(|d_i-\hat{d_i}|-0.5\times \delta), \text{\hspace{0.2em}otherwise} 
	\end{cases}
	\label{eq:huber}
\end{equation}
\noindent\textbf{Self-supervised Learning.} To demonstrate the performance of TriStereoNet on real-world data, and because providing the ground-truth labels for a real dataset is costly, unsupervised learning by a self-supervision loss is utilized. For this purpose, we consider photometric and disparity smoothness losses. More specifically, we reconstruct the reference image by warping the target image with the estimated disparity. We then use a combination of SSIM \cite{wang2004image} and L1 loss as in \cite{godard2017unsupervised} (eq. \ref{eq:LP}) to measure image discrepancy between the reconstructed and the original image.
\begin{equation}
	\mathcal{L}_{p}(I,\hat{I})=\frac{1}{n}\sum_{i} \alpha \frac{1-\text{SSIM}\left(\mathbf{I},  \hat{\mathbf{I}}\right)}{2}+(1-\alpha)\left\|\mathbf{I}-\hat{\mathbf{I}}\right\|
	\label{eq:LP}
\end{equation}
Here, we set $\alpha = 0.85$ and use a SSIM with a $3\times 3$ block filter. $\hat{I}$ is the reconstructed image computed given the disparity map and the baseline. Thus, our multi-baseline setup makes it feasible to consider two photometric losses, which correspond to the reconstructed \emph{left} image with narrow and wide baselines, as in eq. \ref{eq:warp}. Note that for warping the middle image to reconstruct the left image, we need to use $\hat{d}/r$ because of $d_{LR}=r\cdot d_{LM}$. 
\begin{equation}
	\begin{split}
		\hat{I}_L^M = \text{WARP}(I_M,\hat{d}/r), \hspace{0.5cm} \hat{I}_L^R = \text{WARP}(I_R,\hat{d})
	\end{split}
	\label{eq:warp}
\end{equation}
For disparity smoothness loss, we adopt the edge-aware smoothness loss in \cite{heise2013pm} (eq. \ref{eq:smooth}) to encourage the disparity to be locally smooth. This is an L1 loss on the disparity gradients weighted by image gradients.
\begin{equation}
	\mathcal{L}_{s}(\hat{d}, I_L) = \frac{1}{n} \sum_{i} \left | \partial_x \hat{d}   \right | e^{-\left \| \partial_x I_{L} \right \|} + \left | \partial_y \hat{d}   \right | e^{-\left \| \partial_y I_L \right \|}
	\label{eq:smooth}
\end{equation}
\noindent\textbf{Final Loss.}
The final loss is as follows:
\begin{equation}
	\mathcal{L}=\lambda_{d} \cdot \mathcal{L}_{d}+\lambda_{p} \cdot \mathcal{L}_{p} +\lambda_{s} \cdot \mathcal{L}_{s},
\end{equation}
where $(\lambda_{d},\lambda_{p},\lambda_{s})$ are loss coefficients. We formulate this loss for a hybrid \emph{supervised} and \emph{self-supervised} training. Namely, when training on the synthetic dataset, the photometric loss is ignored, \ie $\lambda_{p}=0$, and when the real dataset is used, $\lambda_{d}=0$.

\noindent\textbf{Iterative Sequential Learning Scheme.} Here, we propose a training approach that is highly efficient with no need to ground-truth disparity of the real dataset. To achieve this, we begin by training the model by self-supervised learning on a real dataset with $\mathcal{L}_{s}$ and $\mathcal{L}_{p}$. Then, proceeding with the synthetic dataset, we employ $\mathcal{L}_{s}$ and $\mathcal{L}_{d}$ losses for supervised learning. This way, we optimize the network iteratively with sequential self-supervised and supervised learning. This technique has three advantages: \emph{1)} In the absence of ground-truth disparity, we can train the model on real-world data. \emph{2)} Synthetic dataset supervision assists the network in learning finer details, since considering only image reconstruction loss ($\mathcal{L}_{p}$) yields blurry results. \emph{3)} Using self-supervision without ground-truth helps the network learn the underlying principles of the trinocular setup, preventing it from overfitting to ground-truth labels.
\section{Experiments and Discussion}
\label{ExperimentalResults}
\begin{figure}[t]
	\begin{center}
		\includegraphics[width=1\linewidth]{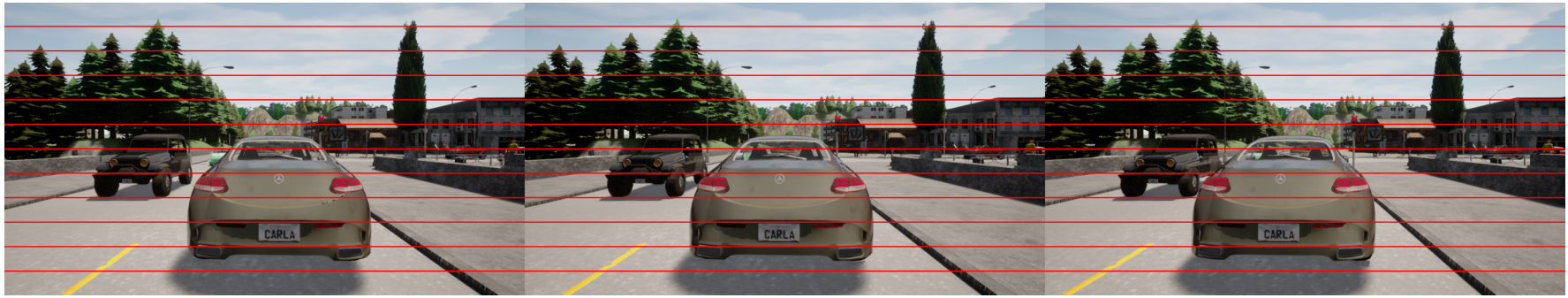}	
		\includegraphics[width=1\linewidth]{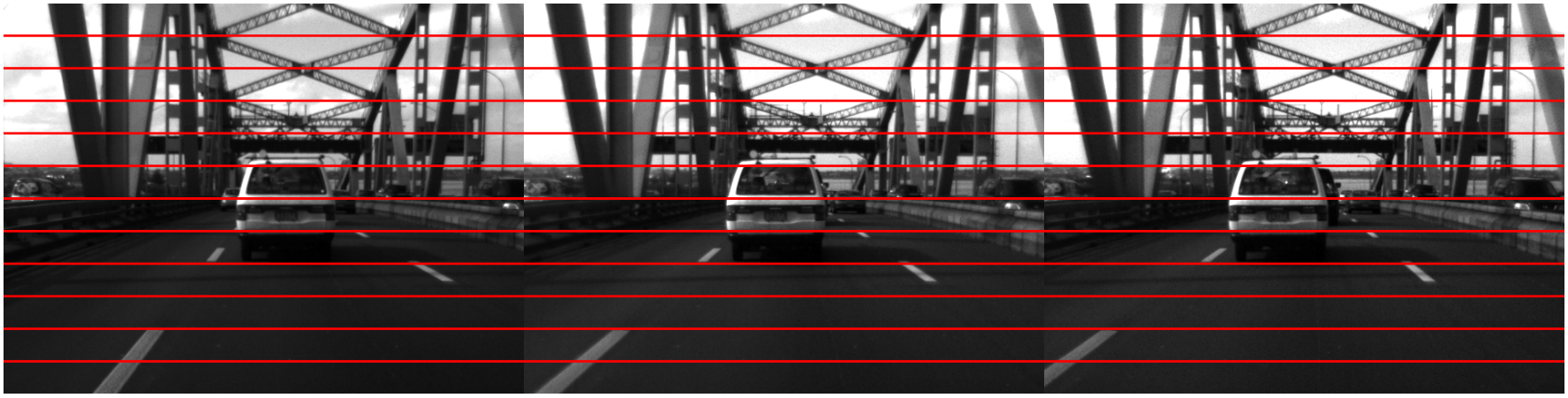}	
		\begin{minipage}[c]{1\linewidth}
			\begin{center}
				\footnotesize{\hspace{0.2cm}Left image \hspace{1.2cm} Middle image \hspace{1.2cm} Right image}	
			\end{center}	
		\end{minipage}
	\end{center}
	\vspace{-0.6cm}
	\caption{\emph{1st Row:} Our synthetic trinocular dataset generated by CARLA. \emph{2nd Row:} The real trinocular images collected by \cite{HER11,SCH11}. As seen through the horizontal lines, the 3-tuple sets in both  real and synthetic datasets are rectified.}
	\label{fig:datasets}
\end{figure}

\noindent\textbf{Synthetic Dataset.} For supervised training of our learning regime, we require the ground-truth data to be available. To the best of our knowledge, there is no publicly available dataset which aims at multi-baseline stereo fusion with in-line cameras and with a high depth range application, like driving. Also, we need a large number of samples for a deep model. Thus, with the help of CARLA \cite{dosovitskiy2017carla}, we generated a synthetic dataset that consists of RGB images of three cameras on an axis with the middle camera centered in the left-right baseline (38.6 $cm$). In order to represent diversified content, we defined 25 configurations based on CARLA's features, such as weather condition, day time, traffic, location, and the number of pedestrians. Accordingly, we created a total number of 9649/2413 training/test samples with images of size $720\times1280$. Note that the images of each triple sample are rectified, and thus, stereo matching and multi-baseline fusion can be performed using any pairs of viewpoints. Figure \ref{fig:datasets} shows a 3-tuple sample of this dataset embedded with horizontal lines.

\noindent\textbf{Real Dataset.} For the real dataset, we use the trinocular set of images collected by \cite{HER11,SCH11}. This dataset is a collection of tricamera stereo sequences with 10 sets as {\tt{Harbour bridge}}, {\tt{Barriers}}, {\tt{Dusk}}, {\tt{Queen street}}, {\tt{People}}, {\tt{Midday}}, {\tt{Night}}, {\tt{Wiper}}, {\tt{Dusk}} and {\tt{Night}}. The last two sets are ignored due to their too bright/dark illumination. The images are 10-bit gray-scale, with a resolution of $480\times640$. Like our CARLA dataset, each 3-tuple sample satisfies the standard epipolar geometry (Fig. \ref{fig:datasets}). We split the dataset into 1920/480 training/test samples. This dataset does not include ground-truth disparity maps.

\noindent\textbf{Evaluation Metrics.} We evaluate the performance of TriStereoNet in terms of:
\begin{itemize}[noitemsep,topsep=0pt,leftmargin=*]
	\item \textbf{EPE} (End-point Error): Mean of absolute error among valid pixels.
	\item \textbf{D1}: Percentage of pixels whose estimation error is $\geq3 px$ or $\geq5 \%$ of the ground-truth disparity.
	\item \textbf{px-1}: Portion of pixels for which absolute error is $\geq1 px$.
	\item \textbf{MRE} (Mean Relative Error) \cite{van2006real}: Mean of absolute error among the valid pixels divided by the ground-truth disparity.
	\item \textbf{px-re-1}: We define this metric as the average of BMPRE metric, defined in \cite{cabezas2012bmpre}, \ie px-re-1 is the percentage of pixels for which MRE is $\geq1$. This metric integrates the benefits of px-1 and MRE. 
\end{itemize}
Note that MRE and px-re-1 consider \emph{depth} (and not disparity). While in recent studies for stereo matching, mainly EPE and D1 measures are evaluated, we believe considering MRE and px-re-1 metrics are equally important as they better represent the actual estimation error in terms of \emph{depth}, which is the ultimate goal of stereo matching. EPE, D1, and px-1 are incompetent in yielding higher error values for larger triangulation errors \cite{van2006real}. 
\begin{figure}[t]
	\begin{center}
		\includegraphics[width=0.6\linewidth]{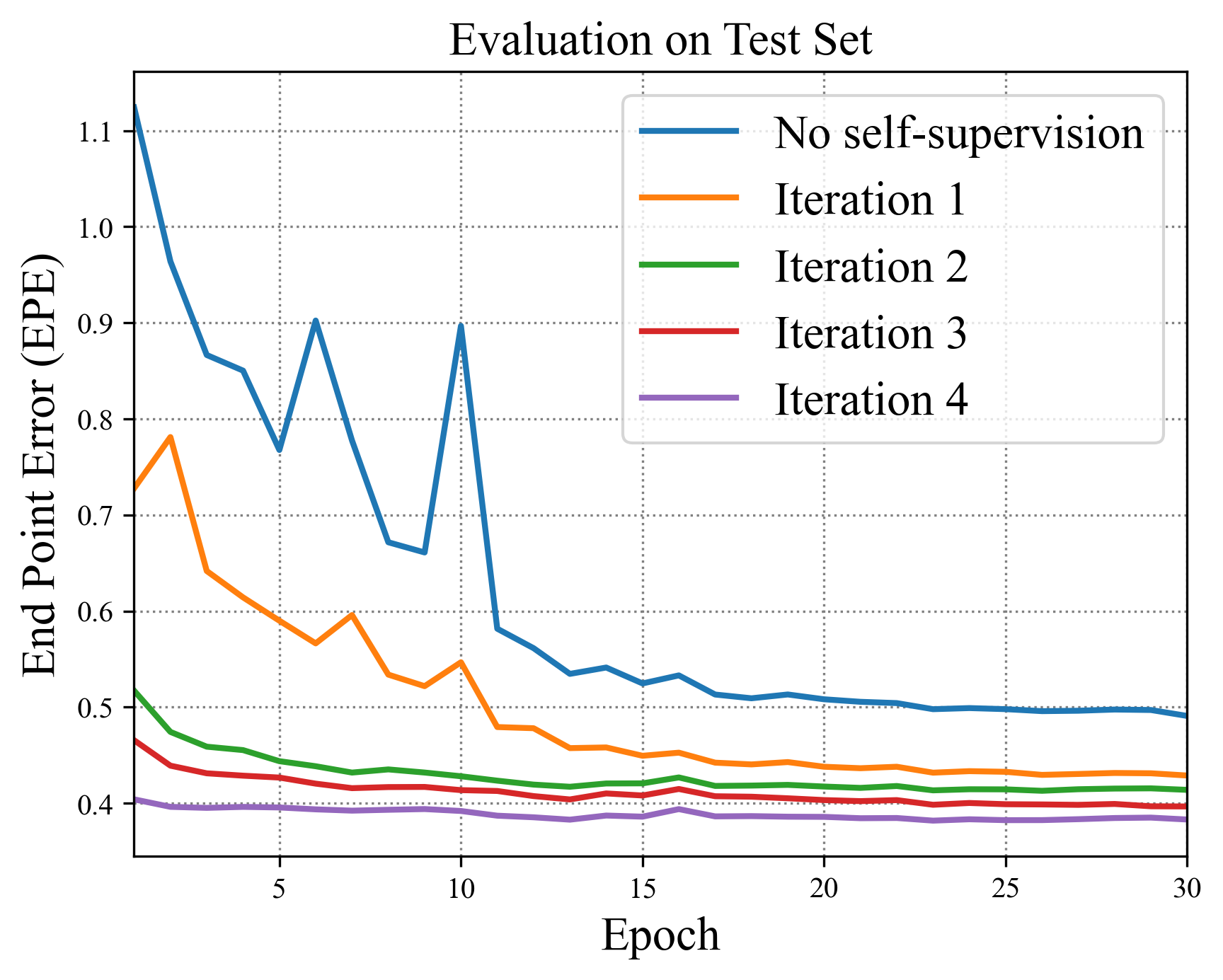}
	\end{center}
	\vspace{-0.8cm}
	\caption{EPE error on the CARLA test set, when the network is trained from scratch with random weights (no self-supervision by the real dataset) and when trained iteratively with self-supervision.}
	\label{fig:learning}
\end{figure}

\noindent\textbf{Implementation Details.} For training, random crops of $256\times512$ are used for both of the datasets. Testing is conducted on crops of $512\times960$ of the synthetic images. For this dataset, the learning rate starts from 0.001 and is downscaled by a factor of 2 after epochs $10,12,14,16$ (in 30 epochs) with Adam optimizer \cite{kingma2014adam}. As for the real dataset, we train the model for 60 epochs with a learning rate downscaled by 2 after epochs $40,50$. We also set $\delta=0.25$ for the Huber loss.
\begin{figure}[tbp]
	\captionsetup[subfigure]{labelformat=empty}
	\centering
	\begin{minipage}[c]{0.985\linewidth}
		\begin{subfigure}[c]{.239\linewidth}
			\includegraphics[width=1\linewidth]{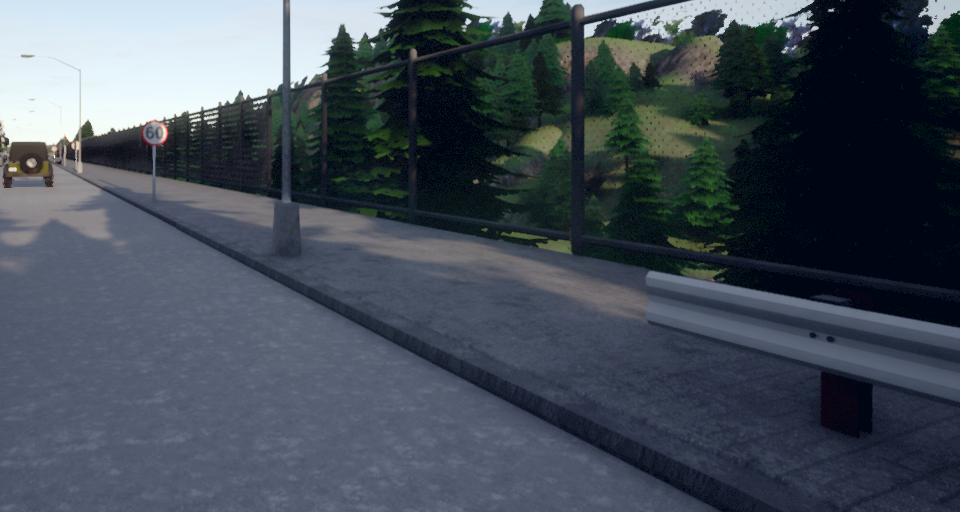}
			\vspace*{-0.38cm}
		\end{subfigure}
		\begin{subfigure}[c]{.239\linewidth}
			\includegraphics[width=1\linewidth]{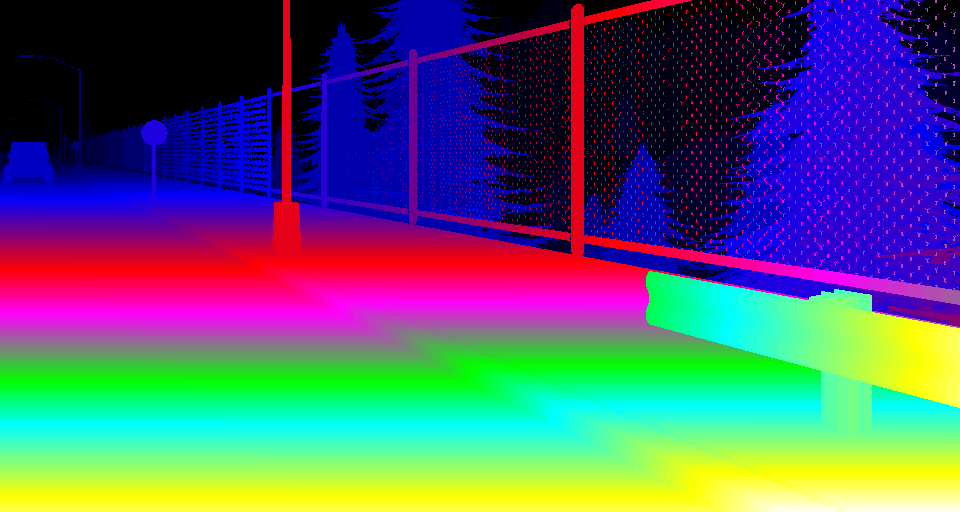}
			\vspace*{-0.38cm}
		\end{subfigure}
		\begin{subfigure}[c]{.239\linewidth}
			\includegraphics[width=1\linewidth]{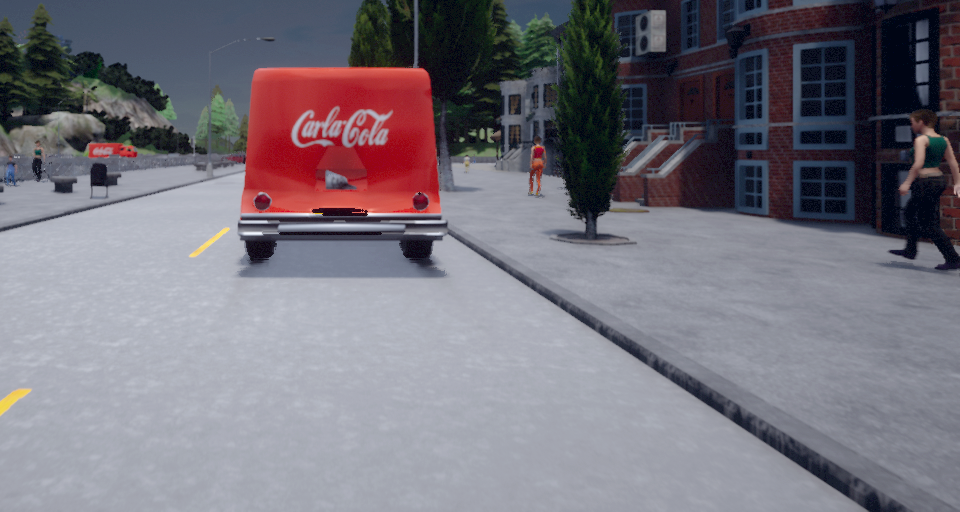}
			\vspace*{-0.38cm}
		\end{subfigure}
		\begin{subfigure}[c]{.239\linewidth}
			\includegraphics[width=1\linewidth]{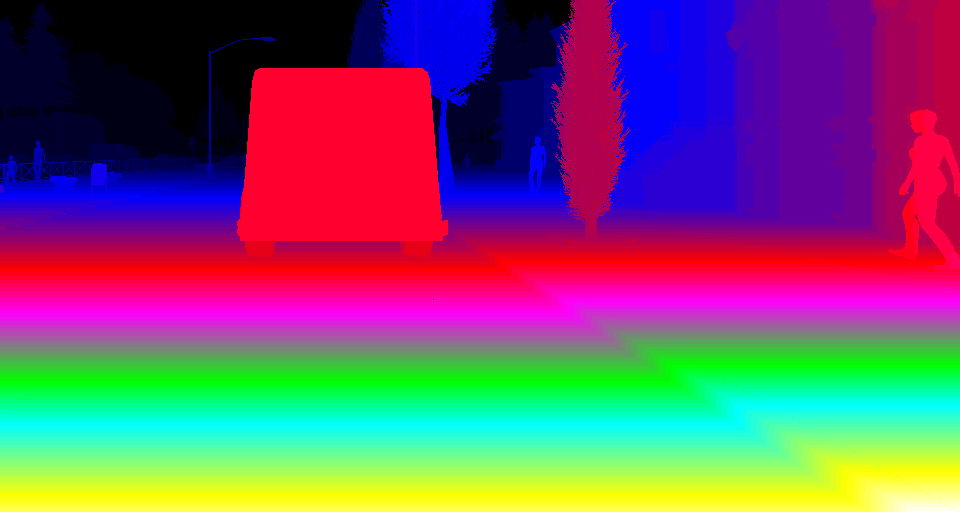}
			\vspace*{-0.38cm}
		\end{subfigure}
	\end{minipage}
	\\
	\begin{minipage}[c]{0.985\linewidth}
		\begin{subfigure}[c]{.239\linewidth}
			\includegraphics[width=1\linewidth]{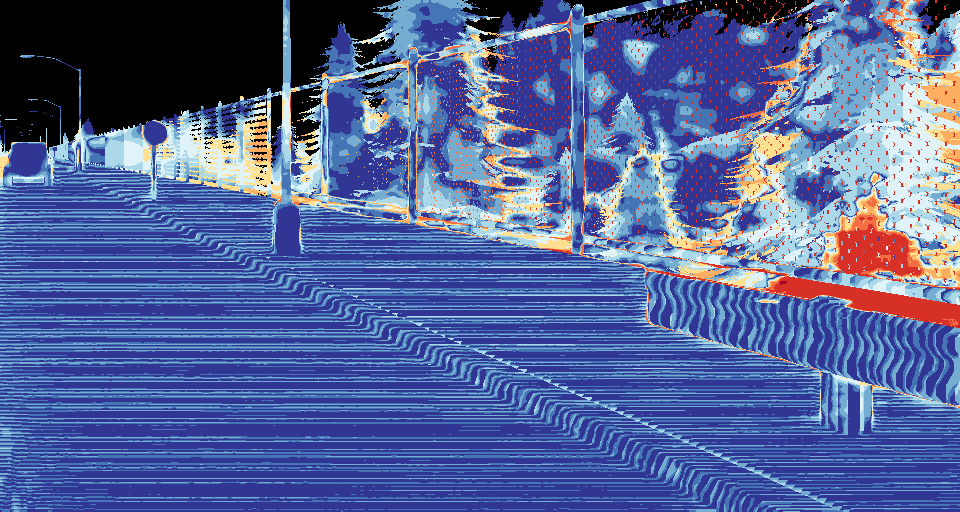}
			\vspace*{-0.38cm}		
		\end{subfigure}
		\begin{subfigure}[c]{.239\linewidth}
			\includegraphics[width=1\linewidth]{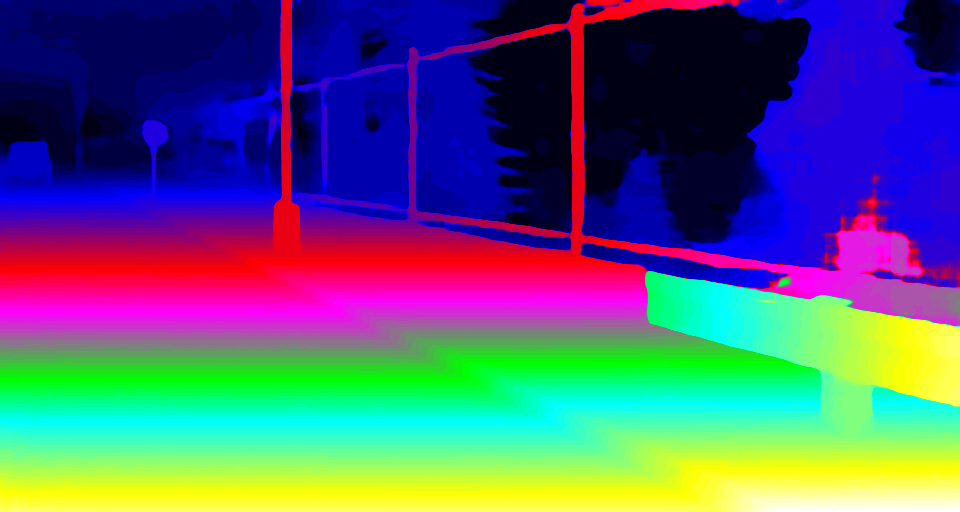}
			\vspace*{-0.38cm}
		\end{subfigure}
		\begin{subfigure}[c]{.239\linewidth}
			\includegraphics[width=1\linewidth]{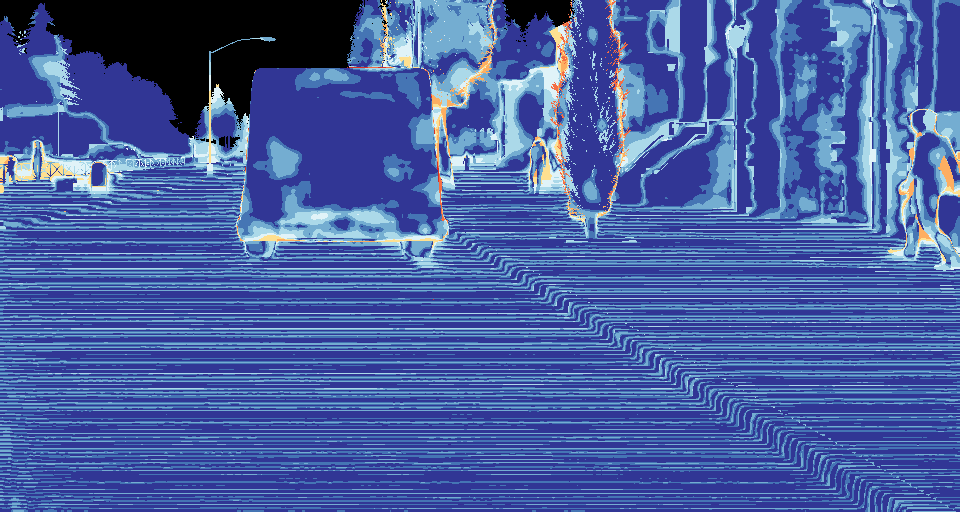}
			\vspace*{-0.38cm}		
		\end{subfigure}
		\begin{subfigure}[c]{.239\linewidth}
			\includegraphics[width=1\linewidth]{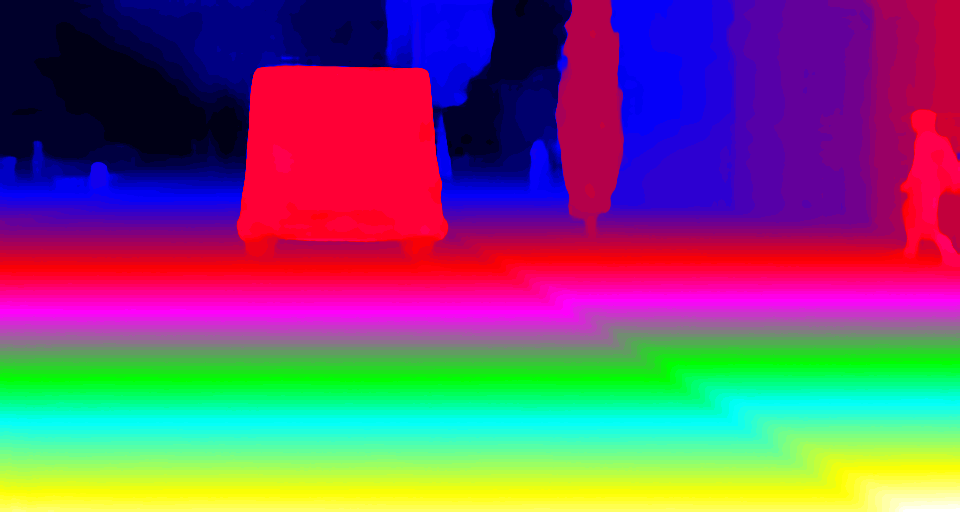}
			\vspace*{-0.38cm}
		\end{subfigure}
	\end{minipage}
	\\
	\begin{minipage}[c]{0.985\linewidth}
		\begin{subfigure}[c]{.239\linewidth}
			\includegraphics[width=1\linewidth]{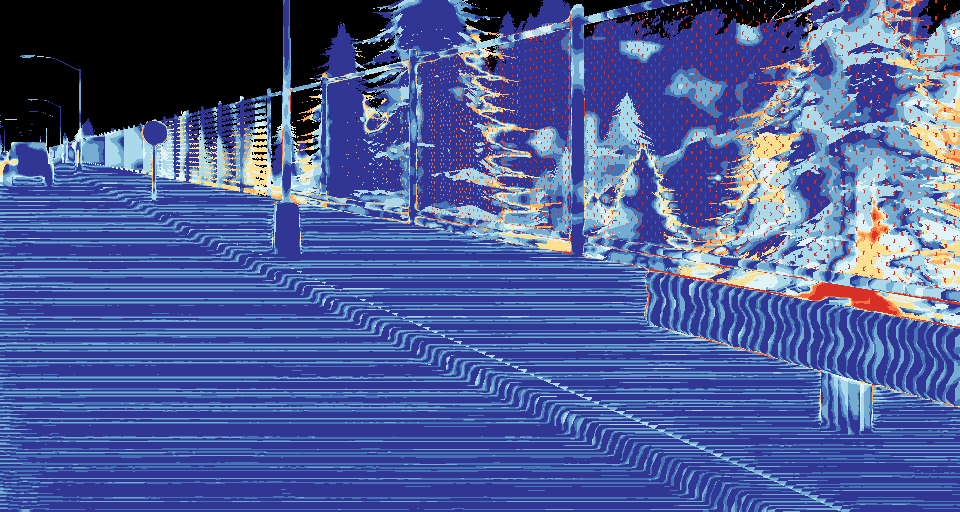}
			\vspace*{-0.38cm}
		\end{subfigure}	
		\begin{subfigure}[c]{.239\linewidth}
			\includegraphics[width=1\linewidth]{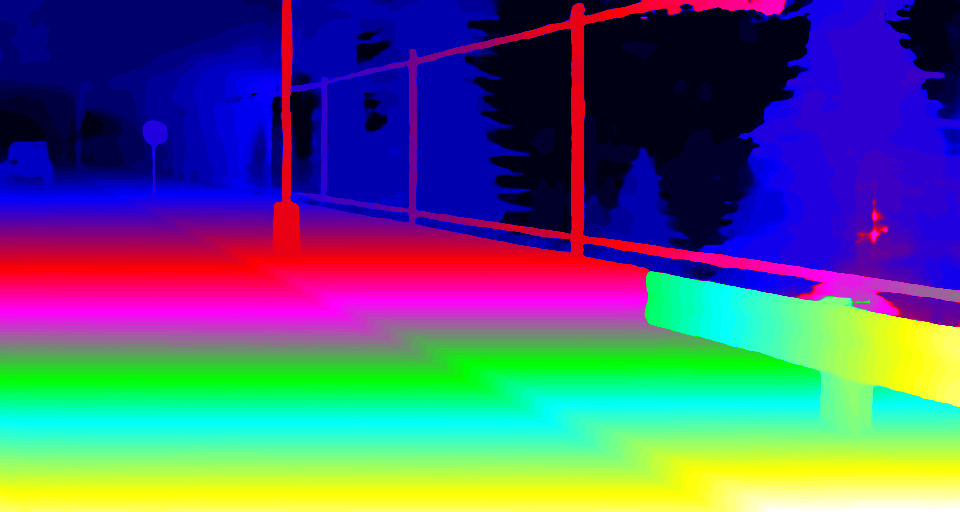}
			\vspace*{-0.38cm}		
		\end{subfigure}
		\begin{subfigure}[c]{.239\linewidth}
			\includegraphics[width=1\linewidth]{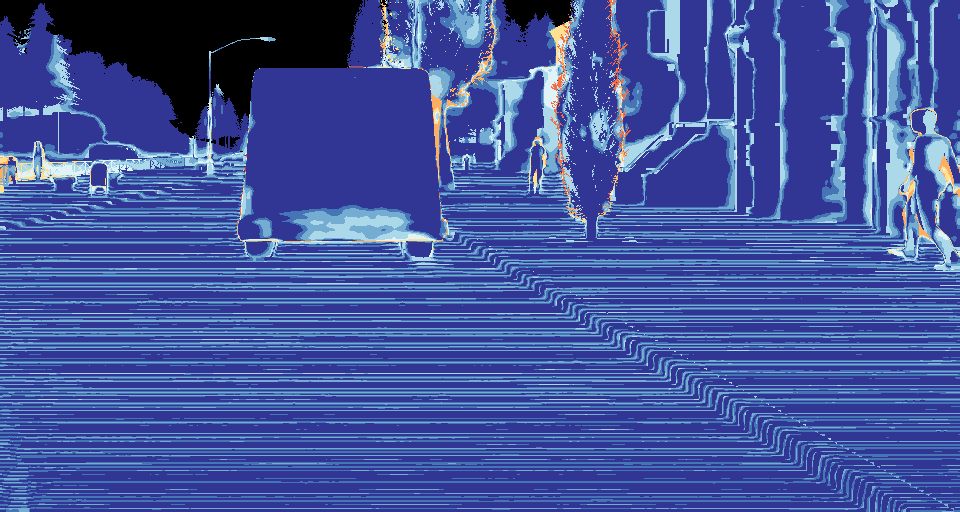}
			\vspace*{-0.38cm}
		\end{subfigure}	
		\begin{subfigure}[c]{.239\linewidth}
			\includegraphics[width=1\linewidth]{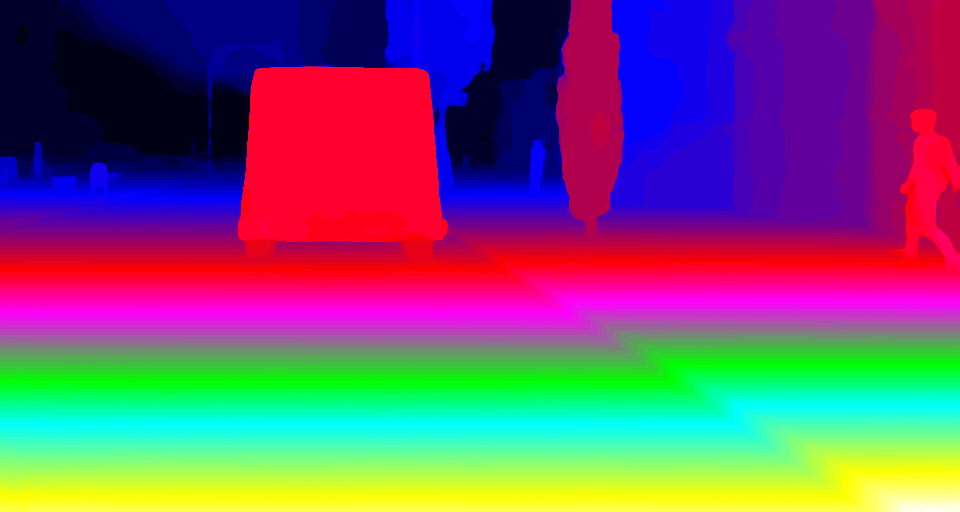}
			\vspace*{-0.38cm}		
		\end{subfigure}
	\end{minipage}	\\
	\vspace{-0.3cm}
	\caption{Disparity estimation and their error maps on the synthetic test set. In the error maps, warmer colors indicate higher error values. \emph{1st Row:} Two input images and their ground-truth disparity map. \emph{2nd Row:} Disparity estimation with vanilla training with no self-supervision. \emph{3rd Row:} Disparity estimation with the proposed sequential learning after four iterations.} 
	\label{fig:final}
\end{figure}
\begin{table}[tbp]
	\begin{center}
		\footnotesize
		\begin{tabular}{@{\hskip1pt}c@{\hskip1pt}|@{\hskip1pt}c@{\hskip1pt}|@{\hskip1pt}c@{\hskip1pt}|@{\hskip1pt}c@{\hskip1pt}|@{\hskip1pt}c@{\hskip1pt}|@{\hskip1pt}c@{\hskip1pt}}
			\hline				
			iteration &    \hspace{0.1cm}EPE($px$) & \hspace{0.1cm}D1(\%) &\hspace{0.1cm}px-1(\%) &  \hspace{0.1cm}MRE & \hspace{0.1cm}px-re-1(\%) \\ \hline
			1 &   0.43  &  1.76  &    5.60   &   0.045   &  0.55  \\
			2 &   0.41  &  1.65  &    5.29   &   0.042   &  0.50  \\ 
			3 &   0.40  &  1.57  &    5.09   &   0.040   &  0.48       \\
			4 &   \textbf{0.38}     &   \textbf{1.49}      &  \textbf{4.86}   &  \textbf{0.038}    &  \textbf{0.44}       \\ \hline
			
		\end{tabular}
	\end{center}
	\vspace{-0.3cm}
	\caption{Performance evaluation on trinocular CARLA dataset. Each iteration consists of sequential self-supervised and supervised training using the real and synthetic datasets. For all metrics, the lower, the better.}
	\label{tab:ietration}
\end{table}

\begin{figure}[tbp]
	\captionsetup[subfigure]{labelformat=empty}
	\centering
	\begin{subfigure}[c]{.325\linewidth}
		\includegraphics[width=1\linewidth]{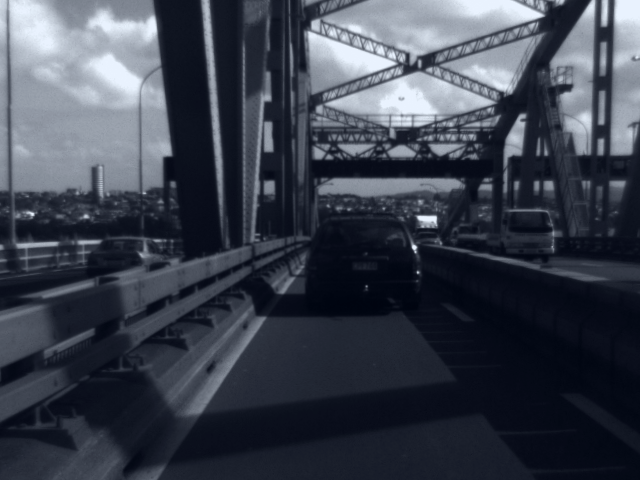}
		\vspace*{-0.38cm}
	\end{subfigure}
	\begin{subfigure}[c]{.325\linewidth}
		\includegraphics[width=1\linewidth]{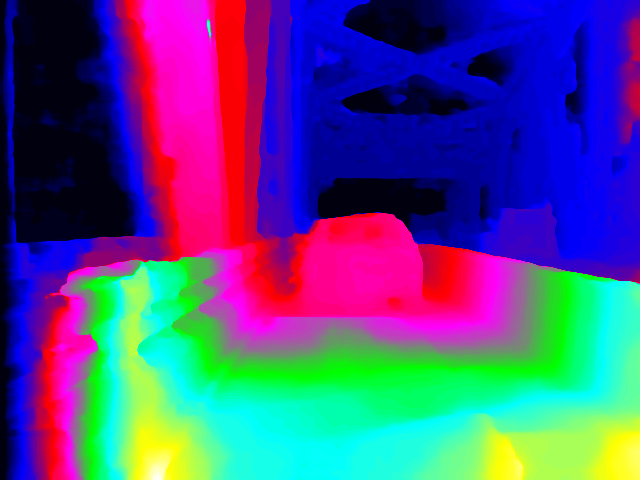}
		\vspace*{-0.38cm}
	\end{subfigure}
	\begin{subfigure}[c]{.325\linewidth}
		\includegraphics[width=1\linewidth]{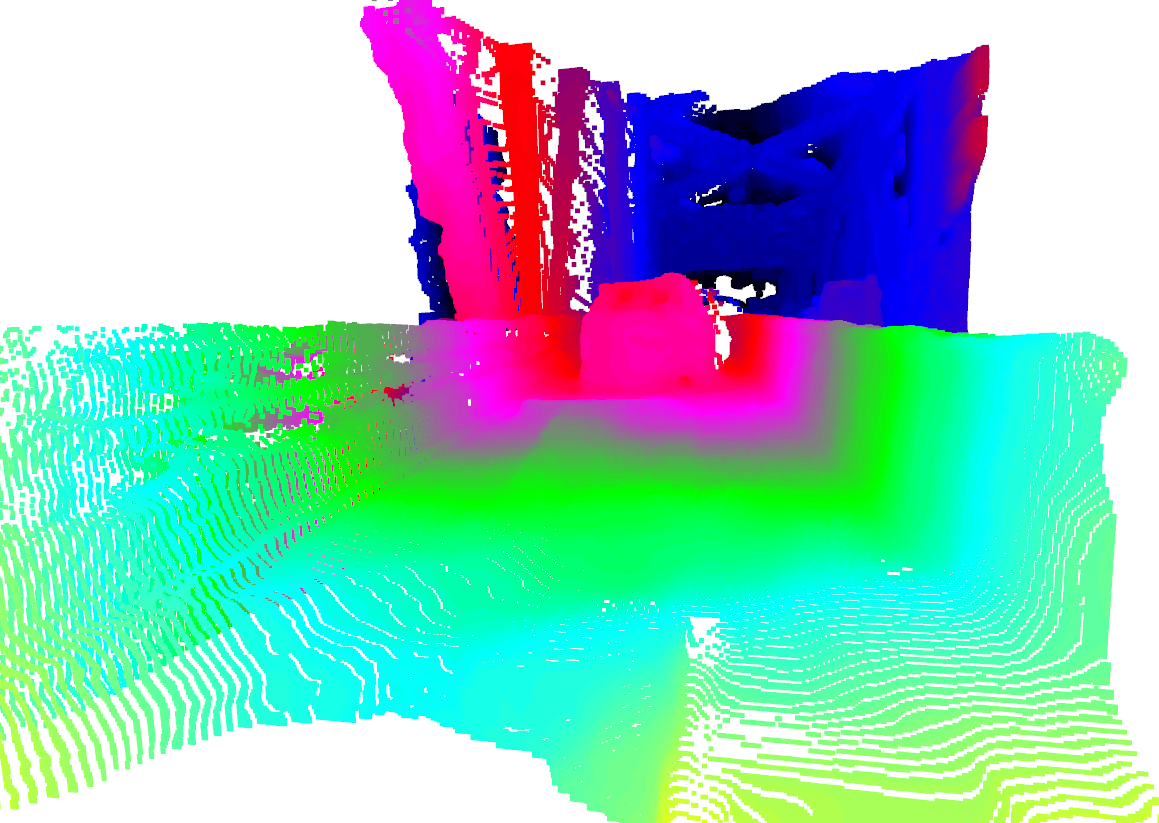}
		\vspace*{-0.38cm}
	\end{subfigure}
	\\
	\begin{subfigure}[c]{.325\linewidth}
		\includegraphics[width=1\linewidth]{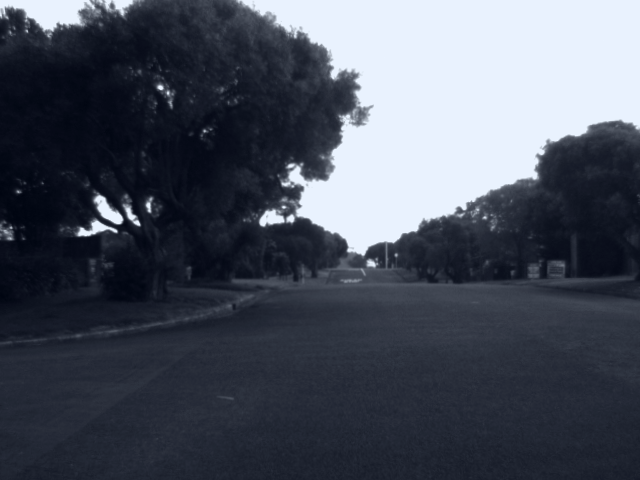}
		\vspace*{-0.38cm}
	\end{subfigure}
	\begin{subfigure}[c]{.325\linewidth}
		\includegraphics[width=1\linewidth]{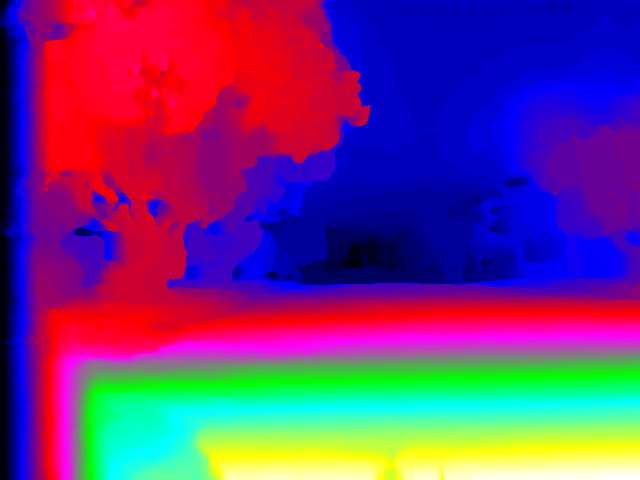}
		\vspace*{-0.38cm}
	\end{subfigure}
	\begin{subfigure}[c]{.325\linewidth}
		\includegraphics[width=1\linewidth]{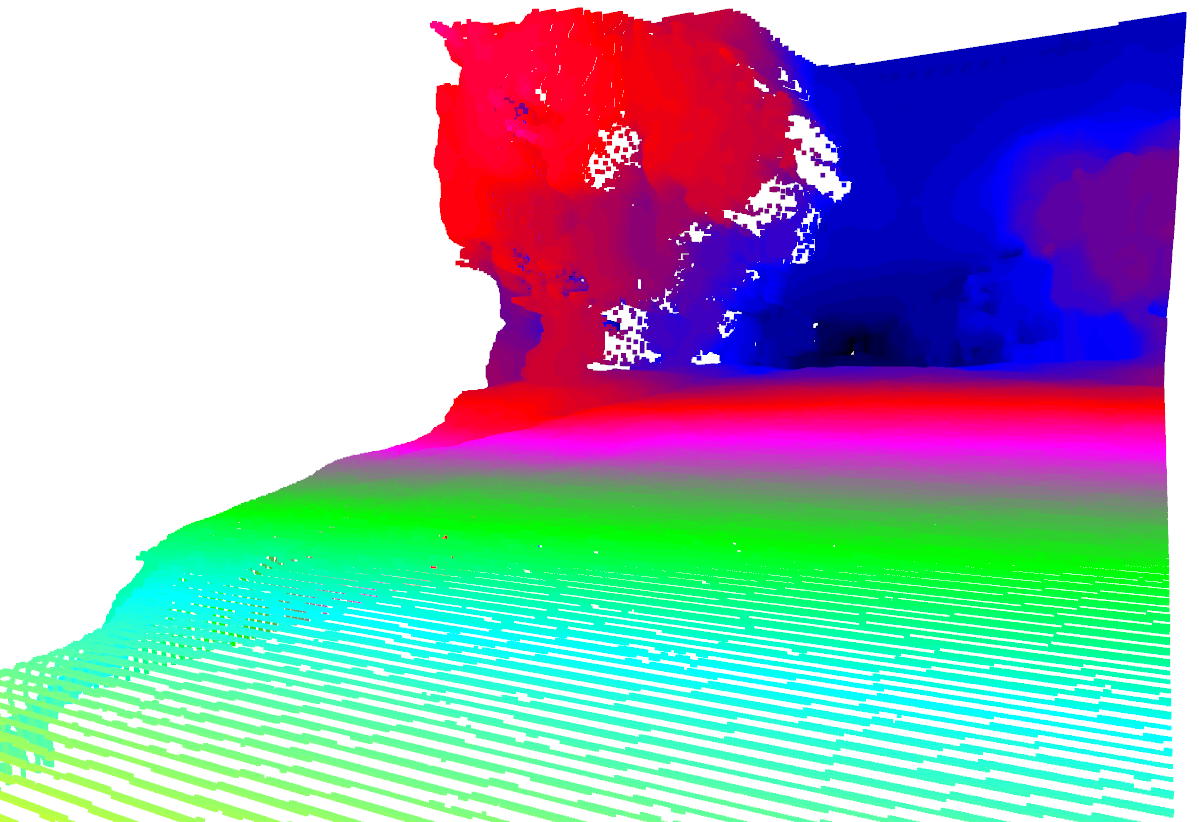}
		\vspace*{-0.38cm}
	\end{subfigure}
	\\
	\begin{subfigure}[c]{.325\linewidth}
		\includegraphics[width=1\linewidth]{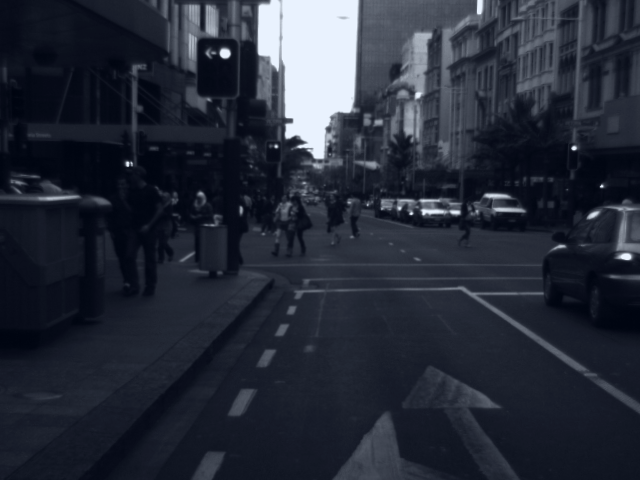}
		\vspace*{-0.45cm}
	\end{subfigure}
	\begin{subfigure}[c]{.325\linewidth}
		\includegraphics[width=1\linewidth]{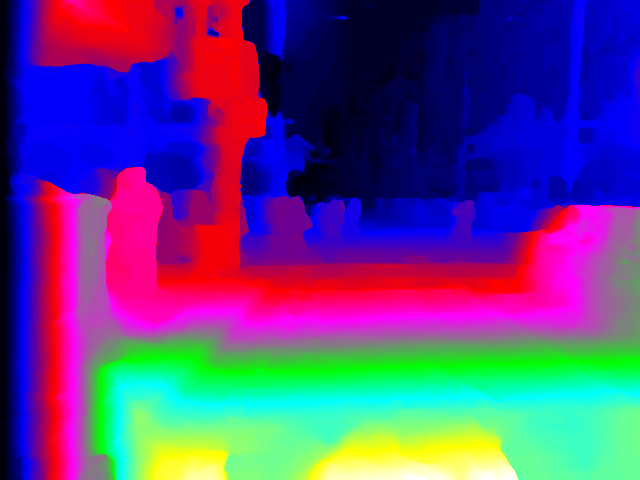}
		\vspace*{-0.45cm}
	\end{subfigure}
	\begin{subfigure}[c]{.325\linewidth}
		\includegraphics[width=1\linewidth]{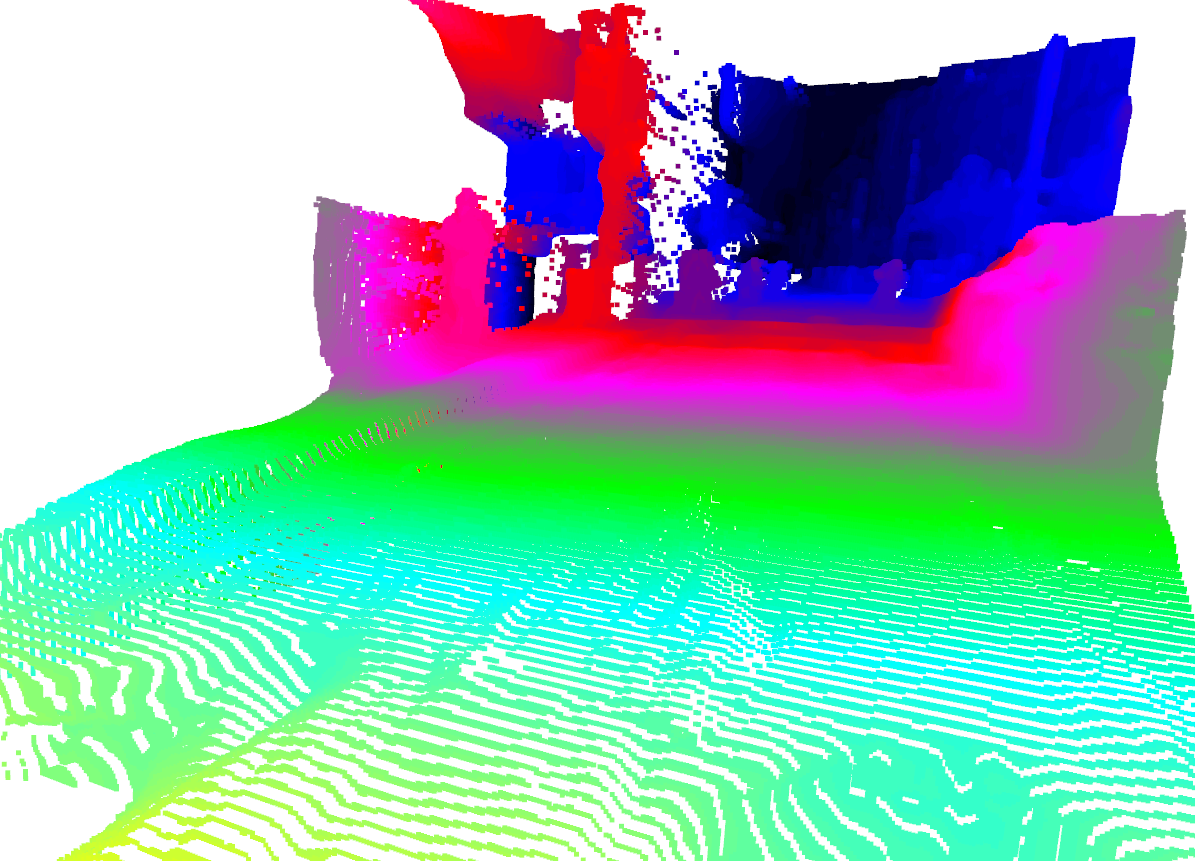}
		\vspace*{-0.45cm}
	\end{subfigure}
	\\
	\vspace{-0.3cm}
	\caption{Disparity estimations on the real dataset. \emph{1st Col.:} Input images (Note that the only similarity of these images with the synthetic dataset is \enquote{being recorded in a driving scenario}). \emph{2nd Col.:} Estimated disparity maps. \emph{3rd Col.:} Reprojectd disparity maps as 3D point clouds.}
	\label{fig:realDataset}
\end{figure}

\noindent\textbf{Quantitative and Qualitative Results.} Table \ref{tab:ietration} presents the evaluation results on the synthetic dataset. We applied iterative sequential learning four times; namely, each iteration started by learning from the real dataset (via self-supervision) and then from the synthetic dataset (via supervised training). From the first iteration to the fourth round, performance improves by approximately 12\%, 15\%, 13\%, 16\%, 20\% reduction in EPE, D1, px-1, MRE, and px-re-1, respectively. With the subsequent iterations, this gain decreases. In Fig. \ref{fig:learning}, the EPE evaluation for different iterations of sequential training is plotted. Note how this learning mechanism surpasses vanilla training, where no self-supervision is carried out. Besides, while the learning process with self-supervision leads to a reasonable convergence in each case, restarting the training with the real dataset further improves the supervised learning. Figure \ref{fig:final} depicts some qualitative results together with the error maps on the synthetic dataset. We can see that the mechanism of iterative sequential training reduces the error. The results on the real dataset together with the 3D reprojections are presented in Fig. \ref{fig:realDataset}. Note that these images are completely different from the synthetic dataset, except that they are both driving scenarios.
\begin{table}[tbp]
	\begin{center}
		\footnotesize
		\begin{tabular}{@{\hskip2pt}l@{\hskip1pt}|@{\hskip2pt}c@{\hskip2pt}c@{\hskip2pt}|@{\hskip2pt}c@{\hskip2pt}c@{\hskip2pt}}
			\hline				
			{Method} & \hspace{0.1cm}{EPE($px$) $\downarrow$} & \hspace{0.1cm}{D1(\%) $\downarrow$} & \hspace{0.1cm}{MACs($G$) $\downarrow$} & \hspace{0.1cm}{Params($M$) $\downarrow$} \\				
			\hline				
			PSMNet\cite{chang2018pyramid} & 0.88 & 2.00  & 256.66 & 5.22 \\
			GA-Ne-deep\cite{zhang2019ga}  & 0.63 & 1.61  & 670.25 & 6.58 \\
			GA-Net-11\cite{zhang2019ga}   & 0.67 & 1.92  & 383.42 & 4.48 \\
			GwcNet-gc\cite{guo2019group}  & 0.63 & 1.55  & 260.49 & 6.82 \\
			GwcNet-g\cite{guo2019group}   & \textbf{0.62} & \textbf{1.49}  & 246.27 & 6.43 \\ \hdashline
			TriStereoNet                  & 0.64 & 1.71 & \textbf{228.30} & \textbf{4.21} \\
			\hline				
		\end{tabular}
	\end{center}
	\vspace*{-0.3cm}
	\caption{Error metrics on the KITTI 2015 validation set and the computational complexity. For TriStereoNet, we assume the middle image is missed and is replaced by the right image. For computing the complexity in terms of MACs, the input resolution is $256 \times 512$.}
	\label{tab:kitti2015}
\end{table}
\subsection{Binocular Deployment}
The performance of TriStereoNet is evaluated on the binocular setup as well. For this, ignoring the viewpoint of the middle camera, we finetune the network on 159/40 training/validation split from the KITTI 2015 dataset \cite{menze2015object}. The results are tabulated in Tab. \ref{tab:kitti2015}, including the error rates as well as the computational complexity in terms of the number of operations (MAC) and parameters. Accordingly, TriStereoNet is also capable of estimating the disparity in binocular setups with similar accuracy but with less complexity than binocular models. Notably, it surpasses GA-Net-11 with 40\% fewer GigaMACs and is competitive with GA-Net-deep with 66\%/36\% fewer operations/parameters. The reason may be due to multi-baseline pre-training and self-supervision, which help the model learn the principles of stereo matching with more constraints from the setup.
\begin{figure*}[htpb]
	\captionsetup[subfigure]{labelformat=empty}
	\centering
	\begin{center}
		\begin{minipage}[c]{0.01\textwidth}
			\begin{turn}{90}\footnotesize{}\end{turn}			
		\end{minipage}
		\begin{minipage}[c]{0.985\textwidth}
			\begin{subfigure}[c]{0.33\linewidth}
				\includegraphics[width=1\linewidth]{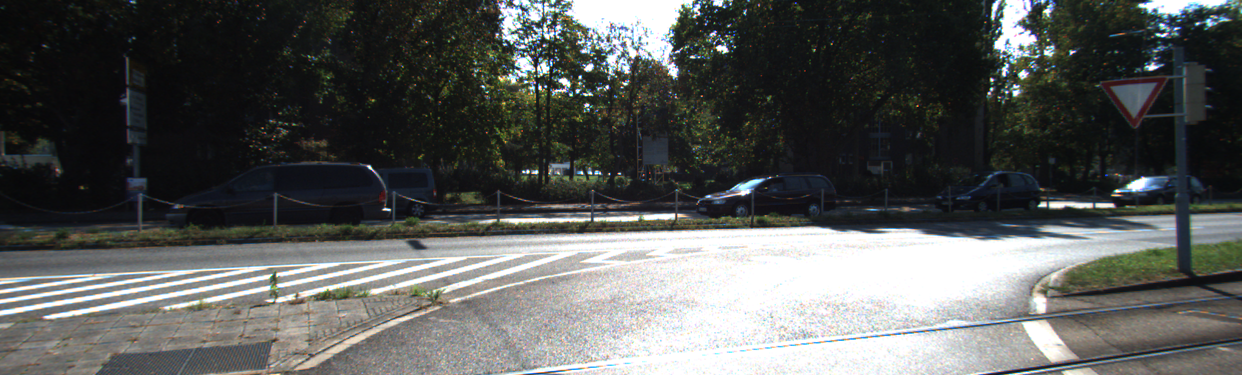}
			\end{subfigure}
			\hspace{-0.5\baselineskip}
			\hfill
			\begin{subfigure}[c]{0.33\linewidth}
				\includegraphics[width=1\linewidth]{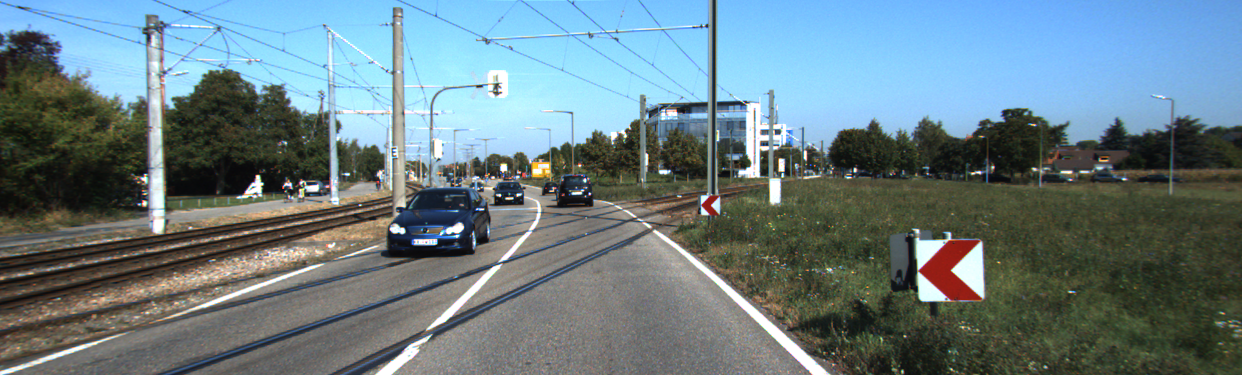}
			\end{subfigure}
			\hspace{-0.5\baselineskip}
			\hfill
			\begin{subfigure}[c]{0.33\linewidth}
				\includegraphics[width=1\linewidth]{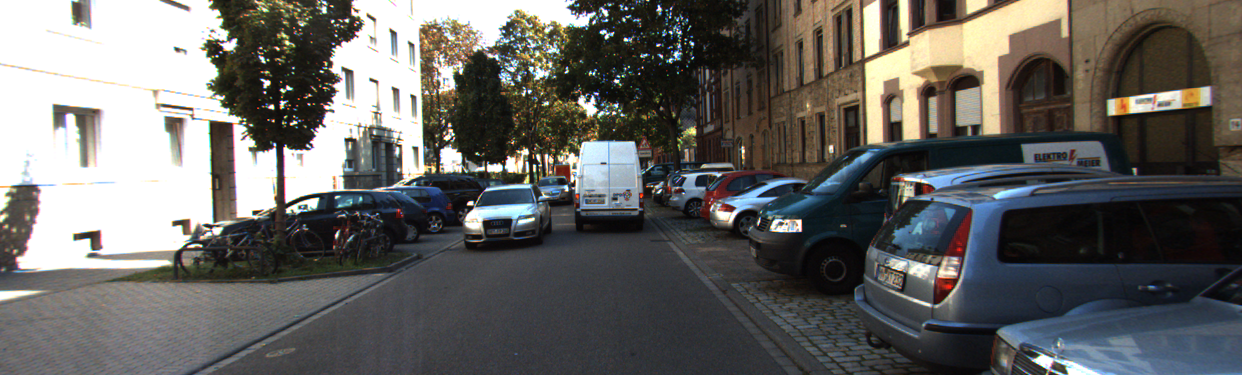}
			\end{subfigure}
		\end{minipage}
		\begin{minipage}[c]{0.01\textwidth}
			\begin{turn}{90}\footnotesize{GCNet \cite{kendall2017end}}\end{turn}			
		\end{minipage}
		\begin{minipage}[c]{0.985\textwidth}
			\begin{subfigure}[c]{0.33\linewidth}
				\includegraphics[width=1\linewidth]{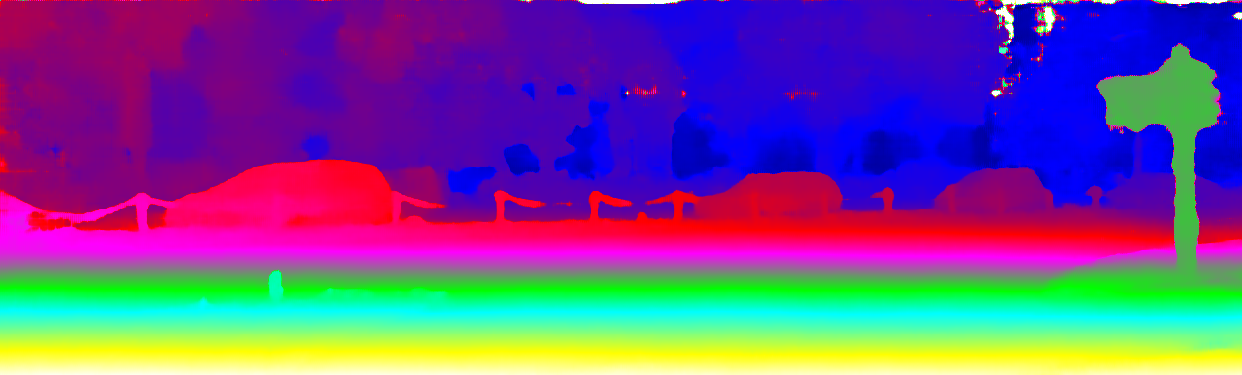}
			\end{subfigure}
			\hspace{-0.5\baselineskip}
			\hfill
			\begin{subfigure}[c]{0.33\linewidth}
				\includegraphics[width=1\linewidth]{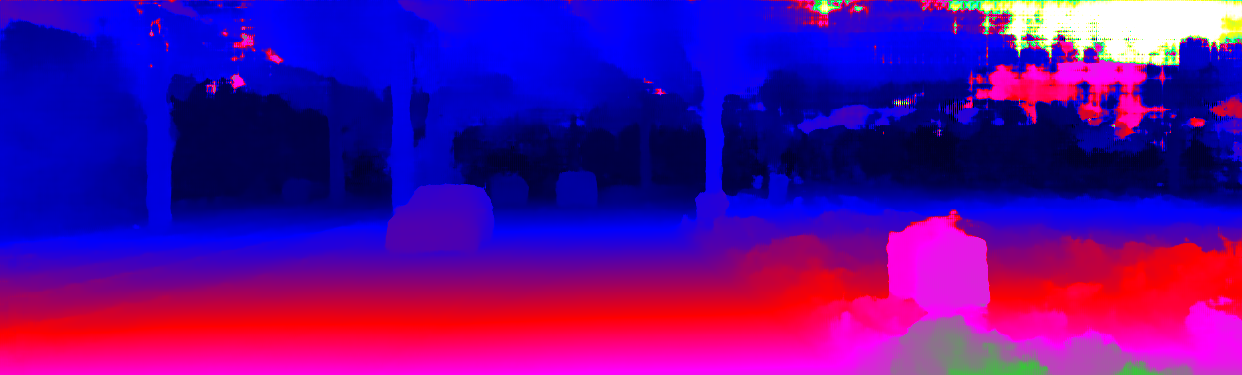}
			\end{subfigure}
			\hspace{-0.5\baselineskip}
			\hfill
			\begin{subfigure}[c]{0.33\linewidth}
				\includegraphics[width=1\linewidth]{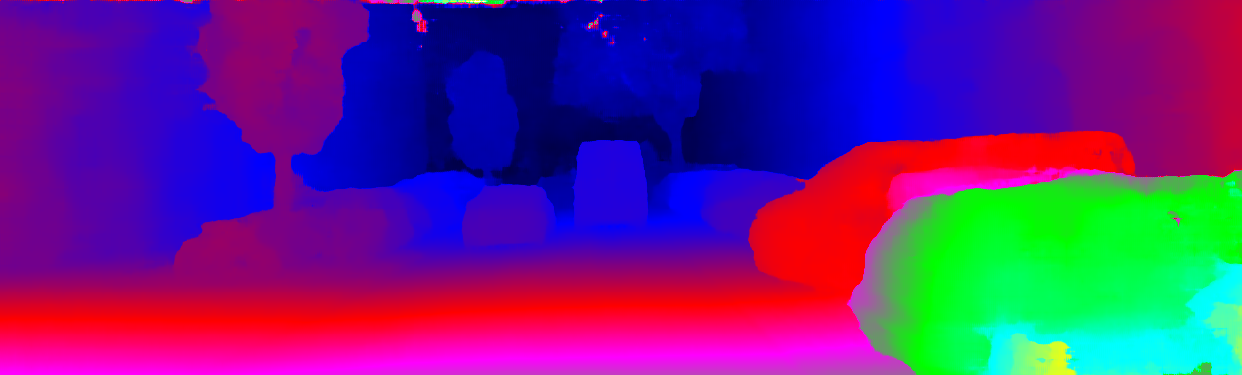}
			\end{subfigure}
			\begin{subfigure}[c]{0.33\linewidth}
				\includegraphics[width=1\linewidth]{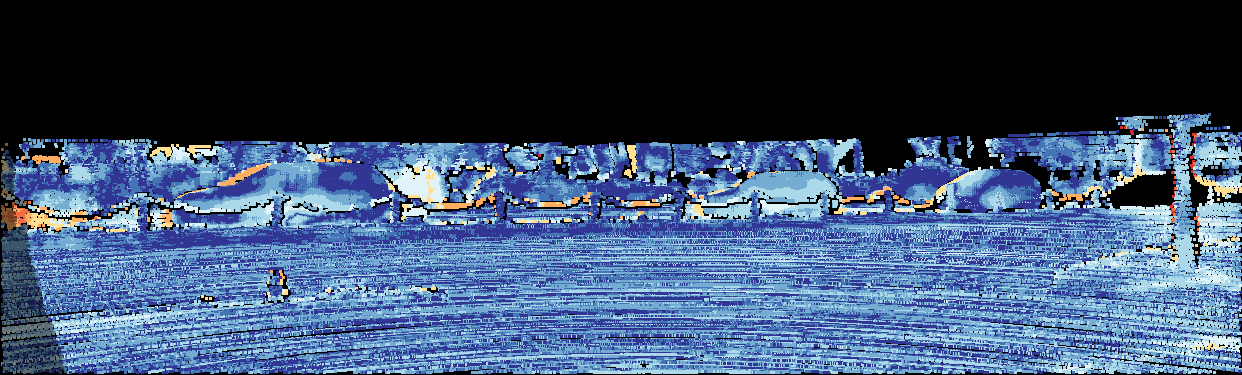}
			\end{subfigure}
			\hspace{-0.5\baselineskip}
			\hfill
			\begin{subfigure}[c]{0.33\linewidth}
				\includegraphics[width=1\linewidth]{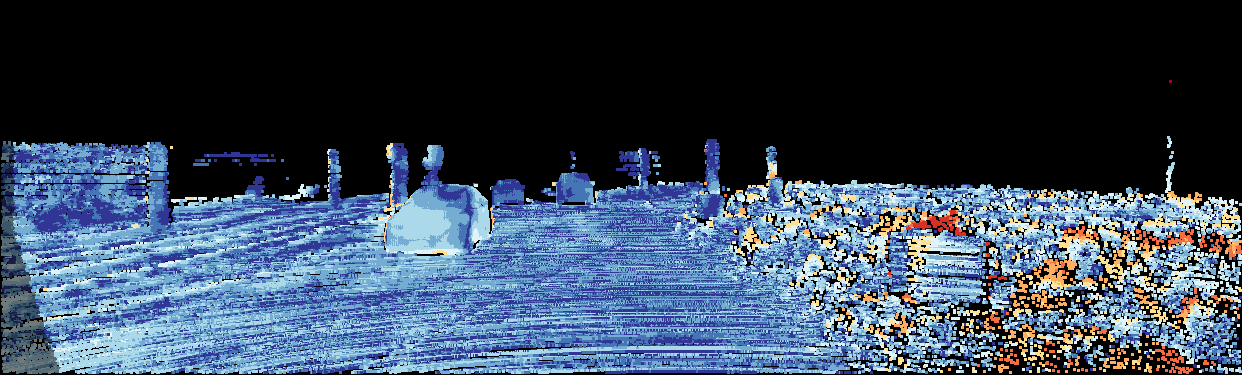}
			\end{subfigure}
			\hspace{-0.5\baselineskip}
			\hfill
			\begin{subfigure}[c]{0.33\linewidth}
				\includegraphics[width=1\linewidth]{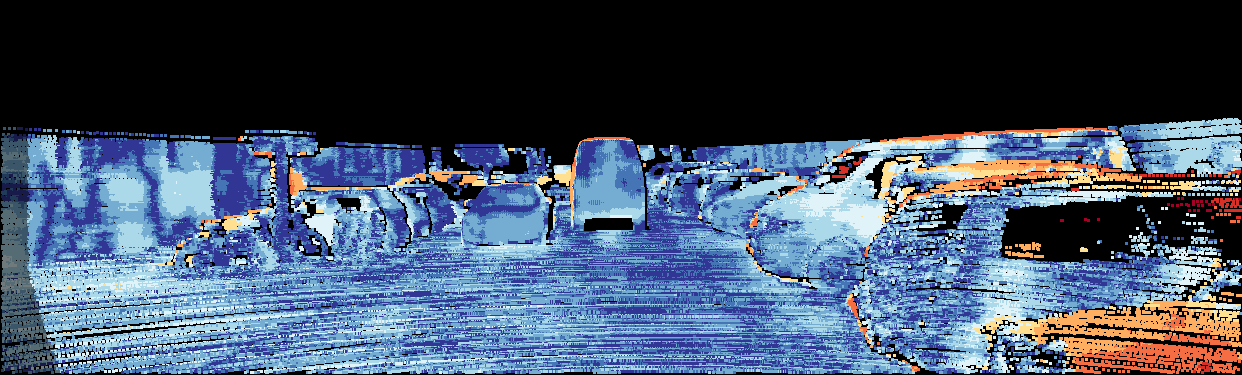}
			\end{subfigure}
		\end{minipage}
		\begin{minipage}[c]{0.01\textwidth}
			\begin{turn}{90}\footnotesize{PSMNet \cite{chang2018pyramid}}\end{turn}			
		\end{minipage}
		\begin{minipage}[c]{0.985\textwidth}
			\begin{subfigure}[c]{0.33\linewidth}
				\includegraphics[width=1\linewidth]{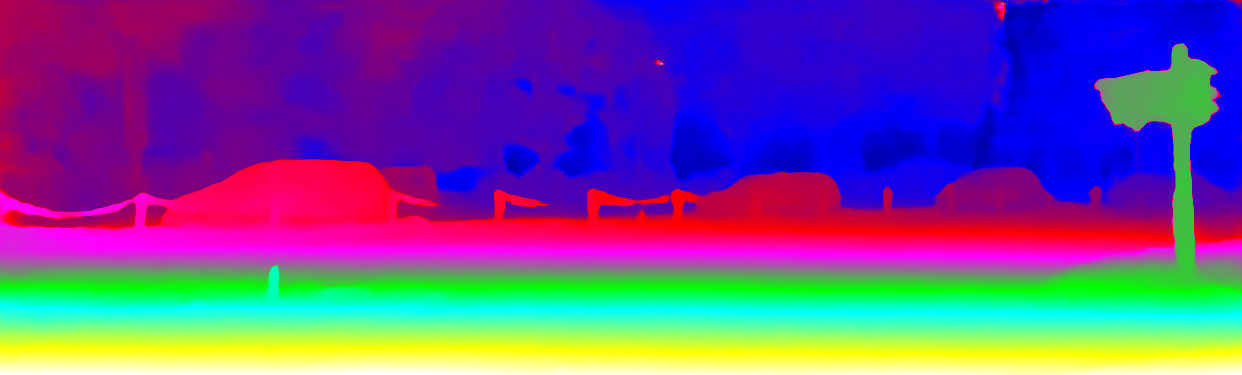}
			\end{subfigure}
			\hspace{-0.5\baselineskip}
			\hfill
			\begin{subfigure}[c]{0.33\linewidth}
				\includegraphics[width=1\linewidth]{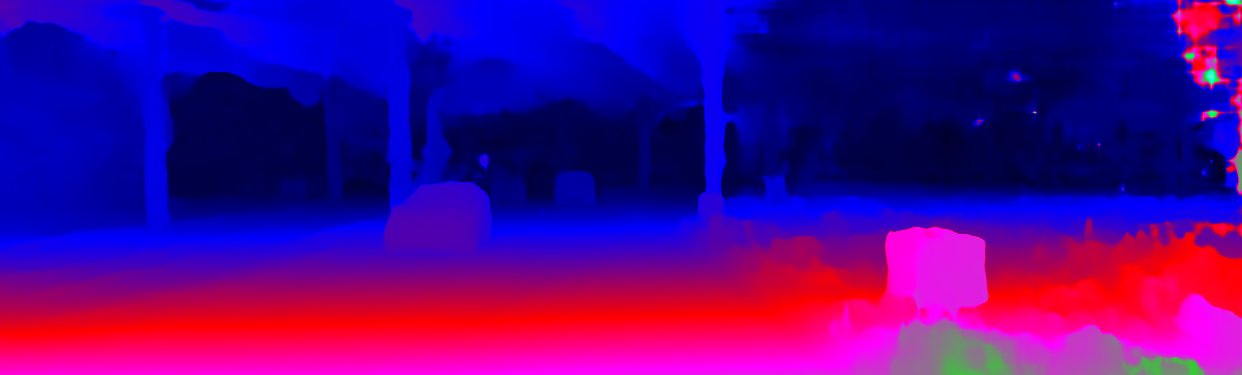}
			\end{subfigure}
			\hspace{-0.5\baselineskip}
			\hfill
			\begin{subfigure}[c]{0.33\linewidth}
				\includegraphics[width=1\linewidth]{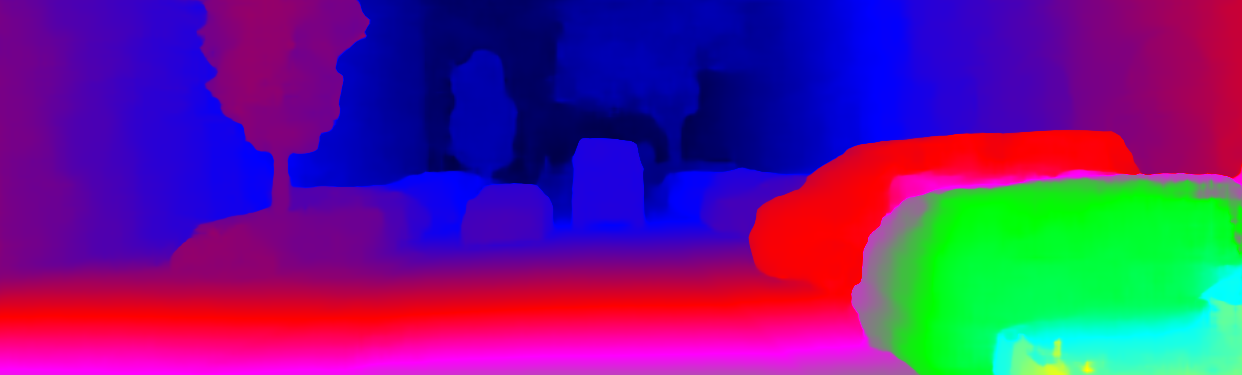}
			\end{subfigure}
			\begin{subfigure}[c]{0.33\linewidth}
				\includegraphics[width=1\linewidth]{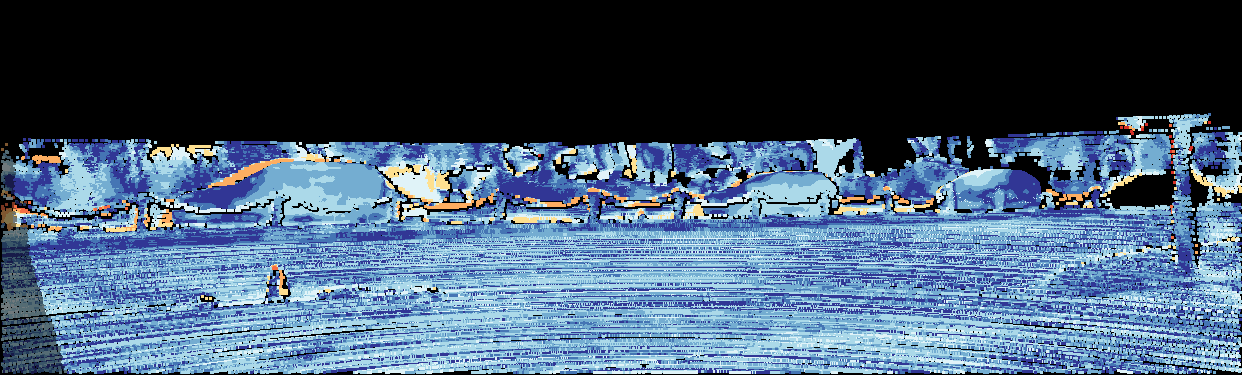}
			\end{subfigure}
			\hspace{-0.5\baselineskip}
			\hfill
			\begin{subfigure}[c]{0.33\linewidth}
				\includegraphics[width=1\linewidth]{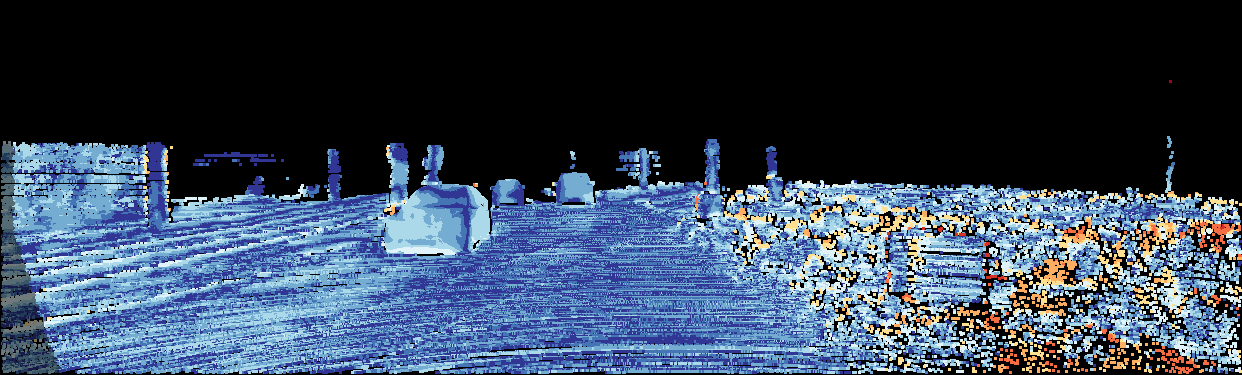}
			\end{subfigure}
			\hspace{-0.5\baselineskip}
			\hfill
			\begin{subfigure}[c]{0.33\linewidth}
				\includegraphics[width=1\linewidth]{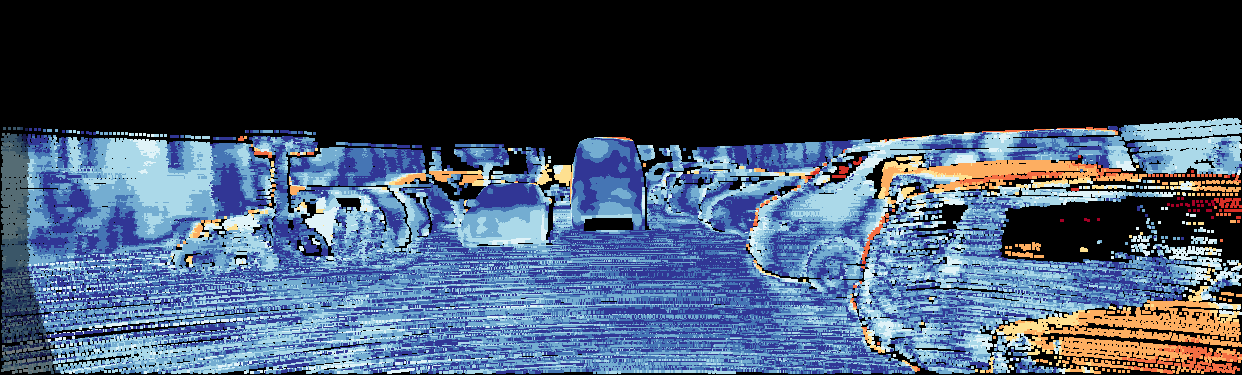}
			\end{subfigure}
		\end{minipage}
		\begin{minipage}[c]{0.01\textwidth}
			\begin{turn}{90}\footnotesize{GwcNet-g \cite{guo2019group}}\end{turn}			
		\end{minipage}
		\begin{minipage}[c]{0.985\textwidth}
			\begin{subfigure}[c]{0.33\linewidth}
				\includegraphics[width=1\linewidth]{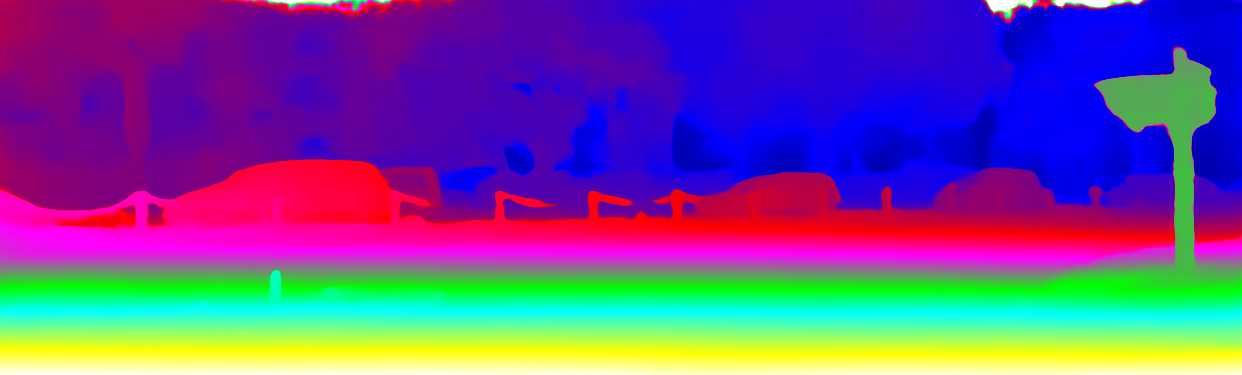}
			\end{subfigure}
			\hspace{-0.5\baselineskip}
			\hfill
			\begin{subfigure}[c]{0.33\linewidth}
				\includegraphics[width=1\linewidth]{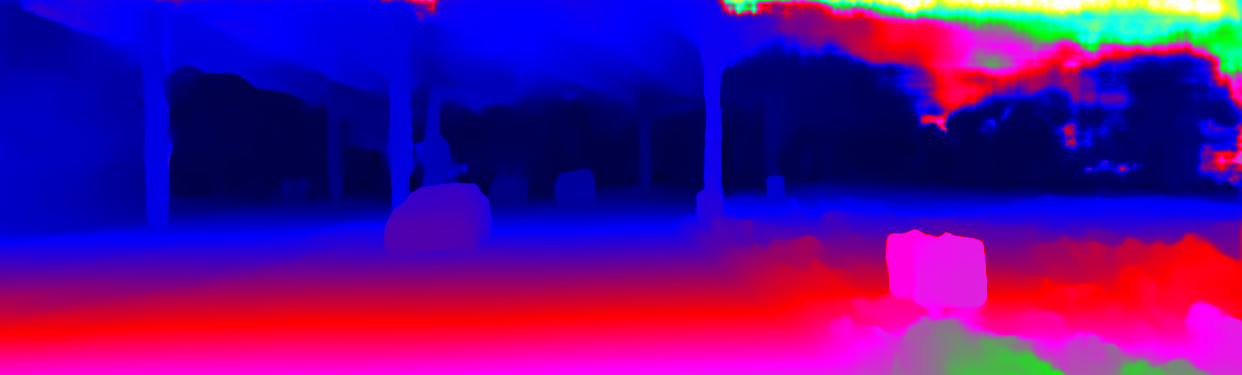}
			\end{subfigure}
			\hspace{-0.5\baselineskip}
			\hfill
			\begin{subfigure}[c]{0.33\linewidth}
				\includegraphics[width=1\linewidth]{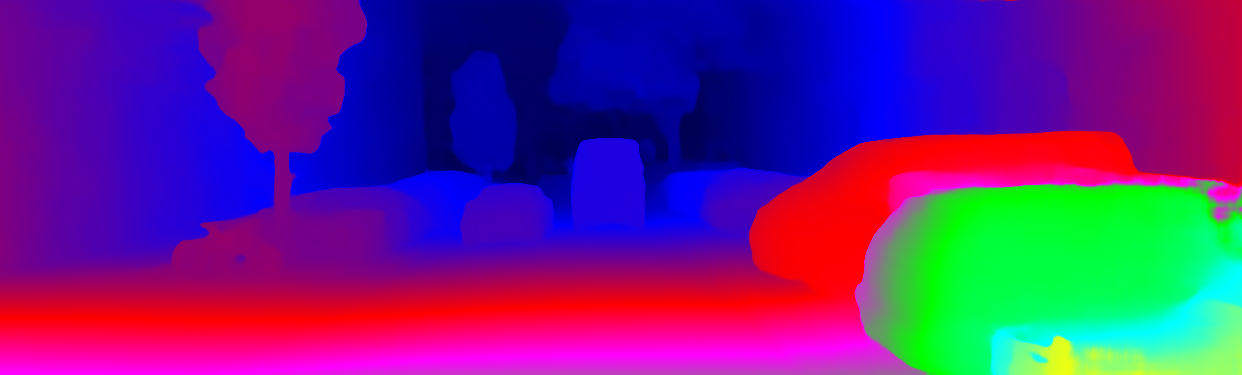}
			\end{subfigure}
			\begin{subfigure}[c]{0.33\linewidth}
				\includegraphics[width=1\linewidth]{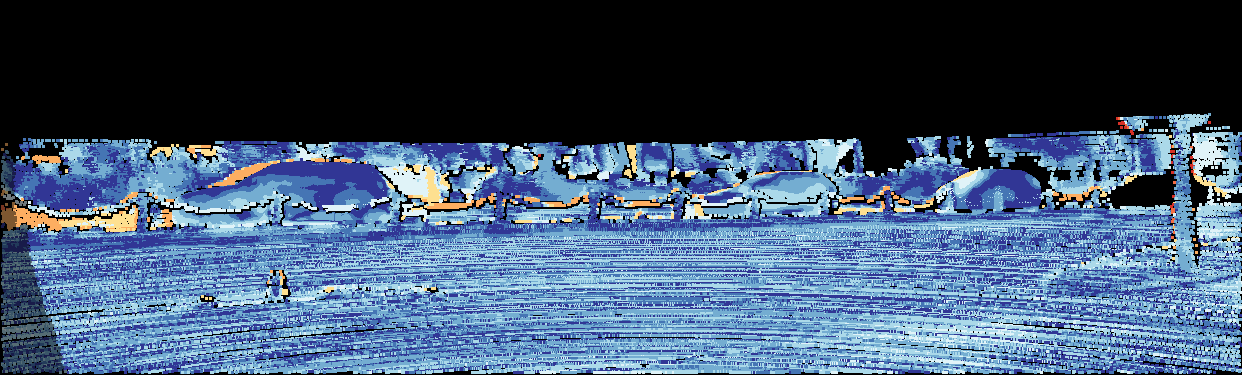}
			\end{subfigure}
			\hspace{-0.5\baselineskip}
			\hfill
			\begin{subfigure}[c]{0.33\linewidth}
				\includegraphics[width=1\linewidth]{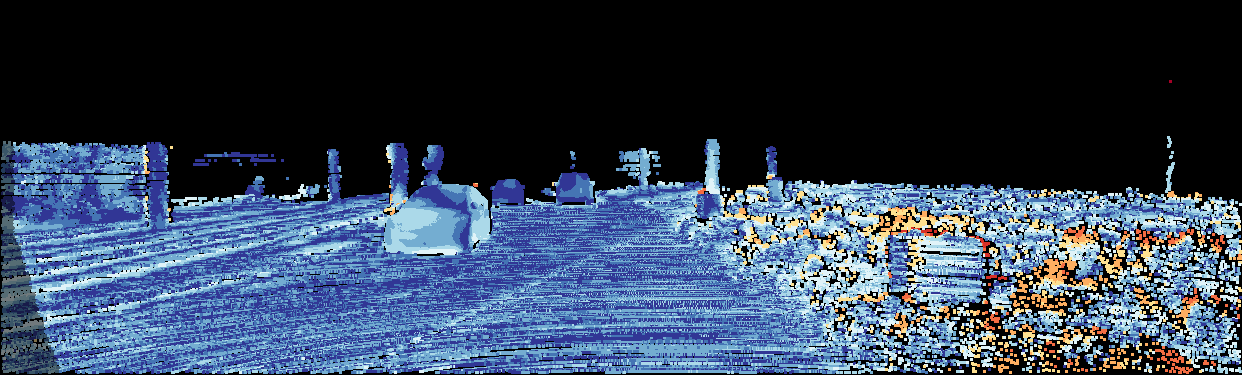}
			\end{subfigure}
			\hspace{-0.5\baselineskip}
			\hfill
			\begin{subfigure}[c]{0.33\linewidth}
				\includegraphics[width=1\linewidth]{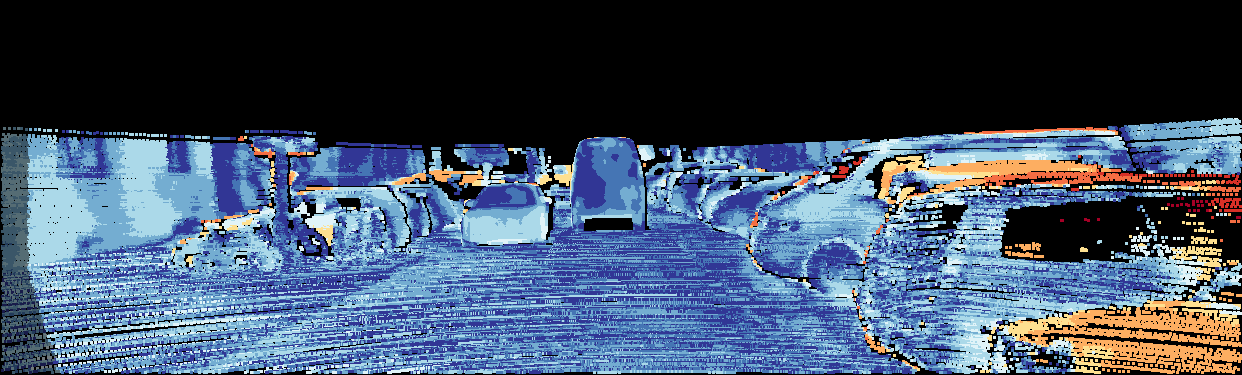}
			\end{subfigure}
		\end{minipage}
		\begin{minipage}[c]{0.01\textwidth}
			\begin{turn}{90}\footnotesize{TriStereoNet}\end{turn}			
		\end{minipage}
		\begin{minipage}[c]{0.985\textwidth}
			\begin{subfigure}[c]{0.33\linewidth}
				\includegraphics[width=1\linewidth]{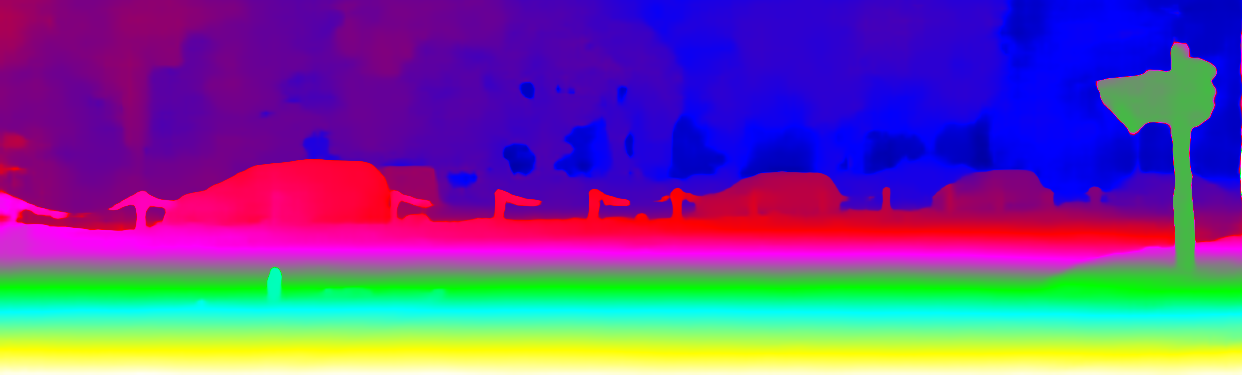}
			\end{subfigure}
			\hspace{-0.5\baselineskip}
			\hfill
			\begin{subfigure}[c]{0.33\linewidth}
				\includegraphics[width=1\linewidth]{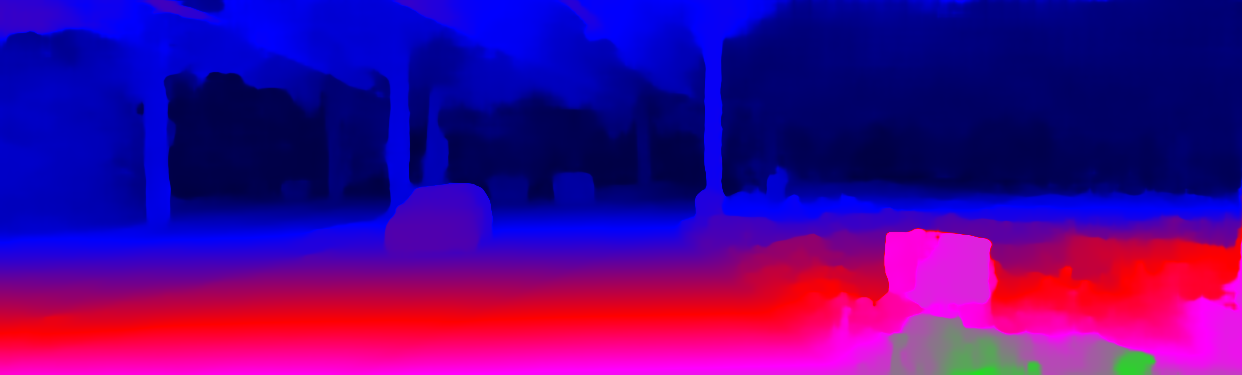}
			\end{subfigure}
			\hspace{-0.5\baselineskip}
			\hfill
			\begin{subfigure}[c]{0.33\linewidth}
				\includegraphics[width=1\linewidth]{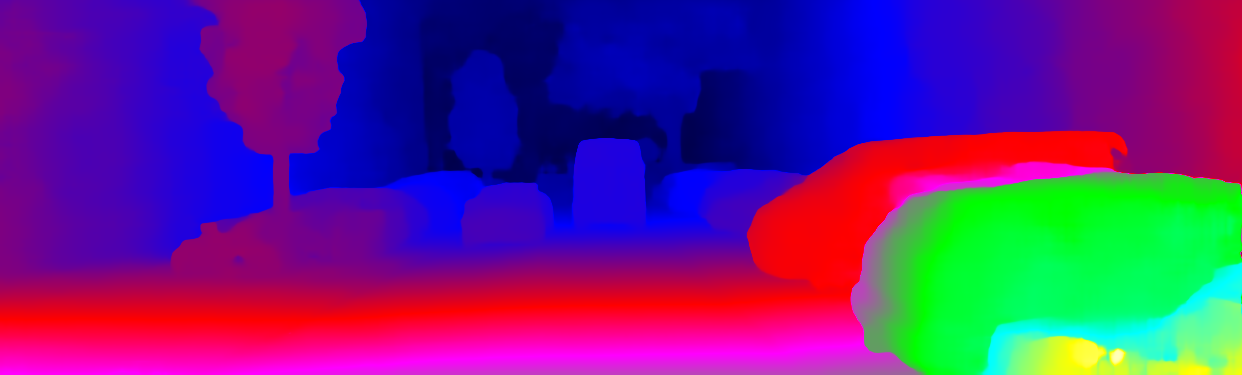}
			\end{subfigure}
			\begin{subfigure}[c]{0.33\linewidth}
				\includegraphics[width=1\linewidth]{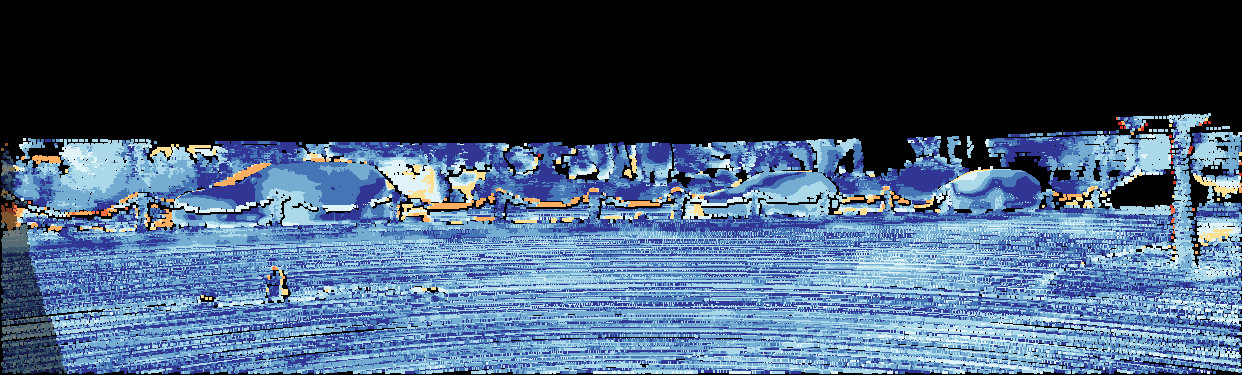}
			\end{subfigure}
			\hspace{-0.5\baselineskip}
			\hfill
			\begin{subfigure}[c]{0.33\linewidth}
				\includegraphics[width=1\linewidth]{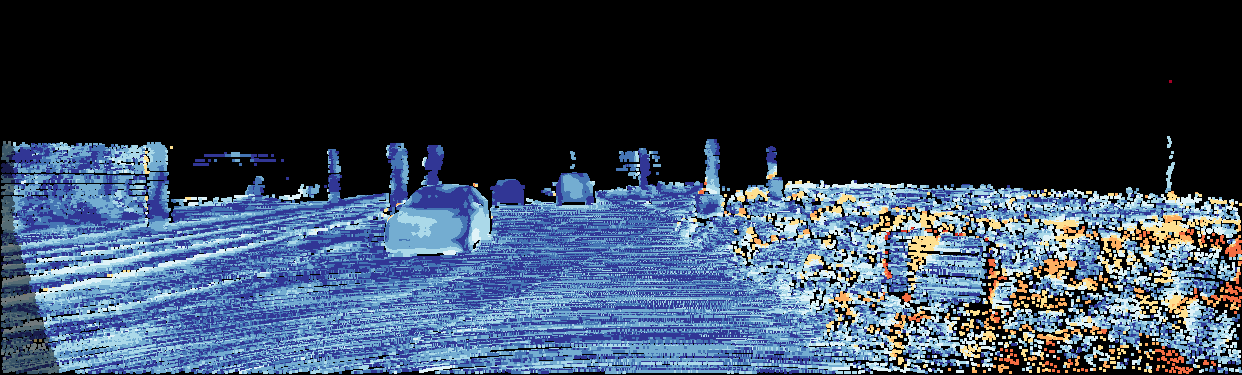}
			\end{subfigure}
			\hspace{-0.5\baselineskip}
			\hfill
			\begin{subfigure}[c]{0.33\linewidth}
				\includegraphics[width=1\linewidth]{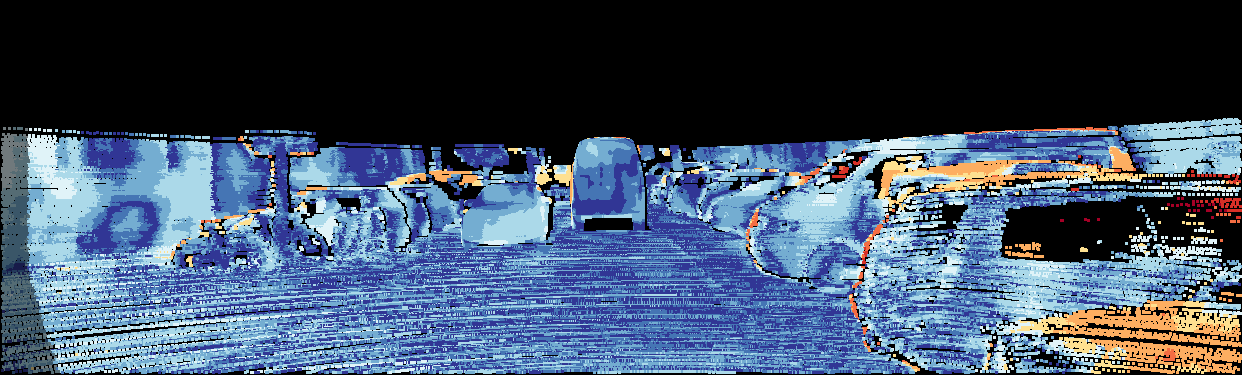}
			\end{subfigure}
		\end{minipage}
		\\
		\vspace{-0.3cm}
		\caption{Qualitative performance on KITTI 2015 benchmark. The first row shows sample left images. The disparity images together with their error maps are estimated by GCNet \cite{kendall2017end}, PSMNet \cite{chang2018pyramid}, GwcNet-g \cite{guo2019group} and TriStereoNet in order.}
		\label{fig:KittiBenchmark}
	\end{center}
\end{figure*}
\begin{figure}[tbp]
	\captionsetup[subfigure]{labelformat=empty}
	\centering
	\begin{subfigure}[c]{.242\linewidth}
		\includegraphics[width=1\linewidth]{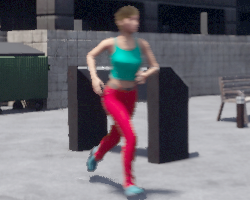}
		\vspace*{-0.38cm}
	\end{subfigure}
	\begin{subfigure}[c]{.242\linewidth}
		\includegraphics[width=1\linewidth]{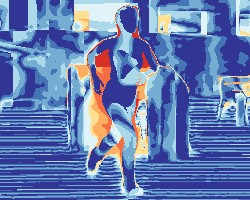}
		\vspace*{-0.38cm}
	\end{subfigure}
	\begin{subfigure}[c]{.242\linewidth}
		\includegraphics[width=1\linewidth]{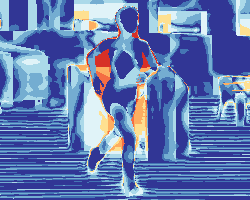}
		\vspace*{-0.38cm}
	\end{subfigure}
	\begin{subfigure}[c]{.242\linewidth}
		\includegraphics[width=1\linewidth]{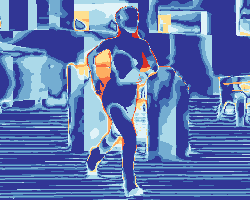}
		\vspace*{-0.38cm}
	\end{subfigure}	
	\\
	\begin{subfigure}[c]{.242\linewidth}
		\includegraphics[width=1\linewidth]{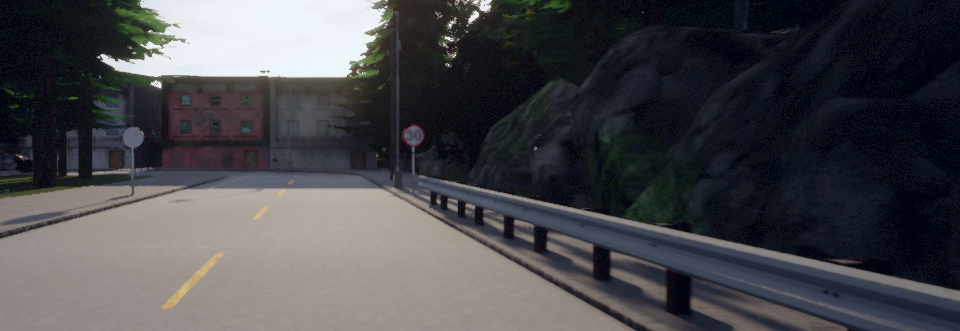}
		\vspace*{-0.45cm}
		\caption{\footnotesize{Left image}}
	\end{subfigure}
	\begin{subfigure}[c]{.242\linewidth}
		\includegraphics[width=1\linewidth]{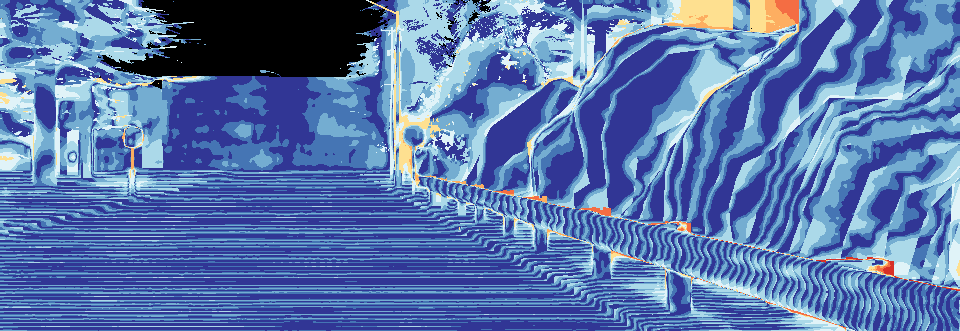}
		\vspace*{-0.45cm}
		\caption{\footnotesize{None}}
	\end{subfigure}
	\begin{subfigure}[c]{.242\linewidth}
		\includegraphics[width=1\linewidth]{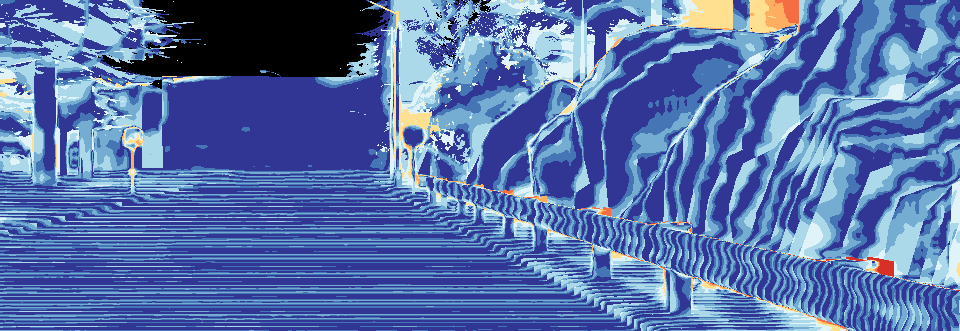}
		\vspace*{-0.45cm}
		\caption{\footnotesize{$\hat{I}_L^M$}}
	\end{subfigure}
	\begin{subfigure}[c]{.242\linewidth}
		\includegraphics[width=1\linewidth]{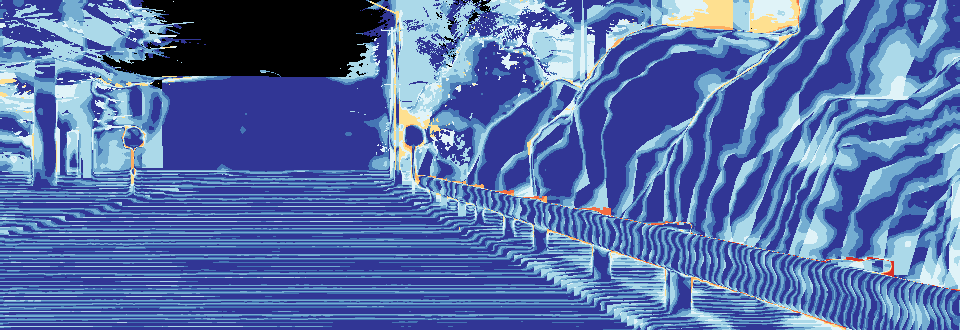}
		\vspace*{-0.45cm}
		\caption{\footnotesize{$\hat{I}_L^R$}}
	\end{subfigure}
	\\
	\vspace{-0.3cm}
	\caption{Disparity error maps for self-supervised initialization (warmer colors denote larger error values). \emph{2nd Column}: the model is trained from scratch on the synthetic dataset. \emph{3rd \& 4th Columns}: self-supervision with the real dataset is performed before the supervised training on the synthetic dataset, using the photometric losses of $\hat{I}_L^M$ and $\hat{I}_L^R$, respectively.}
	\label{fig:warpEffect}
\end{figure}

The results on the KITTI 2015 benchmark (200 test images) are reported in Tab. \ref{tab:kitti}, which also confirm that when TriStereoNet is deployed in a binocular setting, it is still capable of estimating the disparity with superior or competitive accuracy to the inherent binocular models, despite that it has been pretrained with a 3-tuple input. Fig. \ref{fig:KittiBenchmark} illustrates some qualitative comparisons. By visual inspection, we noticed that TriStetreoNet performs more robust in difficult areas, \ie in rich-textured and detailed segments (\eg trees), distant objects/regions (\eg distant cars, trees or sky), and challenging illuminated areas (\eg very bright regions). Note that almost the upper half of the KITTI images do not have the ground-truth annotations of LiDAR. This causes other models to estimate inaccurate or erroneous disparity values in these regions as they cannot learn due to a lack of supervision in those areas. The images' upper half includes trees, buildings, street-covered areas, or the sky. As an example, with other approaches, some sky-related areas are wrongly estimated with very high disparity values, meaning very close objects. TriStereoNet, however, performs with great superiority in these regions as a result of its self-supervised pre-training with three cameras.
\begin{table}[tbp]
	\begin{center}
		\footnotesize
		\begin{tabular}{@{\hskip5pt}l@{\hskip5pt}|@{\hskip5pt}c@{\hskip5pt}c@{\hskip5pt}c@{\hskip5pt}|@{\hskip5pt}c@{\hskip5pt}c@{\hskip5pt}c@{\hskip5pt}}
			\hline
			\multirow{2}{*}{Methods} & \multicolumn{3}{c|}{All(\%)} & \multicolumn{3}{c}{Noc(\%)}  \\	 \cline{2-7} 
			& {D1$_{bg}$} & {D1$_{fg}$} & {D1$_{all}$} & {D1$_{bg}$} & {D1$_{fg}$} & {D1$_{all}$}  \\ \hline
			MC-CNN\cite{zbontar2016stereo} & 2.89 & 8.88 &3.89 & 2.48 & 7.64& 3.33\\
			GCNet\cite{kendall2017end} & 2.21 & 6.16 & 2.87 & 2.02 & 5.58 & 2.61 \\
			AANet \cite{xu2020aanet} & 1.99 & 5.39 & 2.55 & 1.80 & 4.93 & 2.32 \\
			BGNet\cite{xu2021bilateral} & 2.07 & 4.74 & 2.51 & 1.91 & 4.34 & 2.31 \\ 
			MABNet\cite{xing2020mabnet} & 1.89 & 5.02	 & 2.41 & 1.74 & 4.59	 & 2.21	\\ \hdashline	
			\textbf{TriStereoNet} &  1.86 & 4.77 & 2.35 &  1.68 & 4.12 & 2.09  \\\hdashline
			PSMNet\cite{chang2018pyramid} & 1.86 & 4.62 & 2.32 & 1.71 & 4.31 & 2.14 \\
			DeepPruner \cite{duggal2019deeppruner} & 1.87 & 3.56 & 2.15 & 1.71 & 3.18 & 1.95 \\
			GwcNet-g\cite{guo2019group} & 1.74 & 3.93 & 2.11 & 1.61 & 3.49 & 1.92 \\ \hline
		\end{tabular}
	\end{center}
	\vspace*{-0.6cm}
	\caption{Evaluation results on the KITTI 2015 benchmark with D1 measure for non-occluded and all pixels in background, foreground and all areas. The models are sorted according to D1$_{all}$; the lower, the better.}
	\label{tab:kitti}
\end{table}	
\subsection{Ablation Study}
\noindent\textbf{Fusion Level and Fusion Method.} Our first analysis examines the effect of the fusion level (where fusion occurs) and the fusion method for combining narrow- and wide-baseline data. Table \ref{tab:comb} shows the related results when models are only trained on synthetic data. In all of these experiments, the models are trained with 30 epochs, and the best checkpoint is selected based on the least EPE on the test set.

On a cost-level fusion, the Guided Addition layer outperforms the other fusion methods (except for EPE). At this level, we also examined average and simple addition, but discarded them since their results were similar to the single wide baseline. It is possible to achieve a greater gain in performance by applying fusion after both the pre-hourglass and hourglass. As compared with average fusion, we can see that not only the level of fusion is important, but the fusion method is similarly affecting the performance.

The ${hg_{avg}}$ method yields more accurate results in terms of MRE and px-re-1. We believe this is because the hourglass module further processes two data streams. In essence, it works as a regularization step, aggregating disparity information regarding pixels' localities to eliminate unwanted spurious values. Nevertheless, we selected ${pre\_hg_{ga}}$ for merging the data as it involves less computational complexity than in ${hg_{avg}}$. Note that ${hg_{avg}}$ requires processing two data flows in a heavy encoder-decoder with 3D convolutions.

\noindent\textbf{Trinocular \vs Binocular.} To demonstrate the trinocular setup's superior performance over the binocular setup, we exploited a similar network architecture for either the left-middle (LM) or the left-right (LR) image pair. The results are shown in Tab. \ref{tab:comb}. TriStereoNet, which fuses the two baseline streams, outperforms single pairs on all metrics. 

\noindent\textbf{Learning Scheme.} Table \ref{tab:initWarp} presents the impact of self-supervised initialization with the real dataset before training on the synthetic dataset in a supervised manner. It also compares the two methods by which we can reconstruct the reference image. We can see that self-supervised initialization improves the results, particularly when $\hat{I}_L^R$ is used for self-supervision. Figure \ref{fig:warpEffect} shows the qualitative comparison of these approaches.

\noindent\textbf{Disparity Loss.} Last but not least, we evaluated the effect of the Huber loss for comparing disparity maps. Table \ref{tab:Loss} shows that Huber loss is outperforming smooth L1 loss for both binocular and trinocular settings. Note that Huber loss is equivalent to smooth L1 loss when $\delta=1$. Table \ref{tab:Loss} also confirms our choice of $\delta$ as 0.25.
\begin{table}[tbp]
	\begin{center}
		\footnotesize
		\begin{tabular}{@{\hskip1pt}l@{\hskip1pt}|@{\hskip1pt}l@{\hskip1pt}|@{\hskip1pt}c@{\hskip1pt}|@{\hskip1pt}c@{\hskip1pt}|@{\hskip1pt}c@{\hskip1pt}|@{\hskip1pt}c@{\hskip1pt}|@{\hskip1pt}c@{\hskip1pt}}
			\hline				
			& \hspace{0.3cm}Fusion & \hspace{0.1cm}EPE($px$) & \hspace{0.1cm}D1(\%) &\hspace{0.1cm}px-1(\%) &  \hspace{0.1cm}MRE & \hspace{0.1cm}px-re-1(\%) \\ \hline
			LM    &  \hspace{0.5cm} \_    &  0.63  &  2.65       &  8.87   &   0.062   &     0.76   \\  \hdashline
			LR    &  \hspace{0.5cm} \_    &  0.54  &  2.33       &  7.05   &   0.057   &     0.71   \\  \hdashline
			\multirow{10}{*}{LMR} &\hspace{0.1cm}${cost_{max}}$   &  0.53   &   2.21  &  7.02   &  0.055   &  0.66      \\
			&\hspace{0.1cm}${cost_{top}}$   &  0.52   &   2.15  &  6.79   &  0.056   &  0.66      \\ 
			&\hspace{0.1cm}${cost_{cat}}$   &  0.51   &   2.11  &  6.77   &  0.054   &  0.68      \\  
			&\hspace{0.1cm}${cost_{ga}}$    &  0.53   &   2.07  &  6.60   &  0.053   &  0.61      \\ \cdashline{2-7}
			&\hspace{0.1cm}${pre\_hg_{avg}}$ &  0.53  &   2.17  &  6.87   &  0.055   &  0.65     \\
			&\hspace{0.1cm}${pre\_hg_{ga}}$  &  \textbf{0.49}   &   \textbf{2.02}     & 6.39   &    0.052   &   0.64      \\ \cdashline{2-7}
			&\hspace{0.1cm}${hg_{avg}}$     &  0.53        &   2.2      &  7.07    &   \textbf{0.051}  &   \textbf{0.58}     \\
			&\hspace{0.1cm}${hg_{ga}}$      &  \textbf{0.49}        &   \textbf{2.02}      &  \textbf{6.34}    &  0.054   &   0.65      \\  \hline
		\end{tabular}
	\end{center}
	\vspace{-0.3cm}
	\caption{Comparison of the standard stereo with narrow (LM) and wide (LR) baselines (trained with binocular backbone adapted from GwcNet \cite{guo2019group}) and the proposed trinocular stereo (LMR). Furthermore, different \emph{levels} and \emph{methods} of fusion are included. \enquote{${ga}$} indicates the proposed Guided Addition fusion method.}
	\label{tab:comb}
\end{table}

\begin{table}[tbp]
	\begin{center}
		\footnotesize
		\begin{tabular}{@{\hskip1pt}c@{\hskip4pt}c@{\hskip1pt}|@{\hskip1pt}c@{\hskip1pt}|@{\hskip1pt}c@{\hskip1pt}|@{\hskip1pt}c@{\hskip1pt}|@{\hskip1pt}c@{\hskip1pt}|@{\hskip1pt}c@{\hskip1pt}}
			\hline				
			$\hat{I}_L^M$         & $\hat{I}_L^R$        & \hspace{0.1cm}EPE($px$) & \hspace{0.1cm}D1(\%) &\hspace{0.1cm}px-1(\%) &  \hspace{0.1cm}MRE & \hspace{0.1cm}px-re-1(\%) \\ \hline
			&                 &  0.49  &  2.02  &    6.39   &    0.052   &   0.64         \\ \hdashline
			\checkmark    &                 &  0.44  &  1.79  &    5.65   &    0.046   &   0.57       \\
			&   \checkmark    &  \textbf{0.43}  &  \textbf{1.76}  &    \textbf{5.60}   &    \textbf{0.045}   &   \textbf{0.55}    \\ \hline
		\end{tabular}
	\end{center}
	\vspace{-0.3cm}
	\caption{Effect of self-supervised initialization with the real dataset and different photometric losses using $\hat{I}_L^M$ and $\hat{I}_L^R$. While in the first experiment, the model has been trained from scratch, the other two are trained after the self-supervised pre-training.}
	\label{tab:initWarp}
\end{table}
\begin{table}[tbp]
	\begin{center}
		\footnotesize
		\begin{tabular}{@{\hskip1pt}l@{\hskip1pt}|@{\hskip1pt}c@{\hskip1pt}c@{\hskip1pt}|@{\hskip1pt}c@{\hskip1pt}c@{\hskip1pt}|@{\hskip1pt}c@{\hskip1pt}c@{\hskip1pt}}		
			\hline				
			\multicolumn{1}{@{\hskip1pt}c@{\hskip1pt}|}{\multirow{2}{*}{$\delta$}} & \multicolumn{2}{@{\hskip1pt}c@{\hskip1pt}|}{LM}        & \multicolumn{2}{@{\hskip1pt}c@{\hskip1pt}|}{LR}       & \multicolumn{2}{@{\hskip1pt}c@{\hskip1pt}}{LMR}      \\ \cline{2-7} 
			
			\multicolumn{1}{c|}{} & \multicolumn{1}{@{\hskip1pt}c@{\hskip1pt}|}{EPE($px$)} & MRE & \multicolumn{1}{@{\hskip1pt}c@{\hskip1pt}|}{EPE($px$)} & MRE & \multicolumn{1}{@{\hskip1pt}c@{\hskip1pt}|}{EPE($px$)} & MRE \\ \hline
			
			\multicolumn{1}{l|}{1.00} & \multicolumn{1}{c|}{0.68}    &  \hspace{0.1cm}0.067  & \multicolumn{1}{c|}{0.58}    & \hspace{0.1cm}0.061   & \multicolumn{1}{c|}{0.54}    &  \hspace{0.1cm}0.057  \\ 	
			
			\multicolumn{1}{l|}{0.75} & \multicolumn{1}{c|}{0.64}    &  \hspace{0.1cm}0.065  & \multicolumn{1}{c|}{0.57}    & \hspace{0.1cm}0.059   & \multicolumn{1}{c|}{0.53}    &  \hspace{0.1cm}0.057  \\ 	
			
			\multicolumn{1}{l|}{0.50} & \multicolumn{1}{c|}{0.66}    &  \hspace{0.1cm}0.065  & \multicolumn{1}{c|}{0.56}    & \hspace{0.1cm}0.058   & \multicolumn{1}{c|}{0.51}    &  \hspace{0.1cm}0.056  \\ 	
			
			\multicolumn{1}{l|}{0.25} & \multicolumn{1}{c|}{\textbf{0.63}}    & \hspace{0.1cm}\textbf{0.062}   & \multicolumn{1}{c|}{\textbf{0.54}}    & \hspace{0.1cm}\textbf{0.057}   & \multicolumn{1}{c|}{\textbf{0.49}}    &  \hspace{0.1cm}\textbf{0.052}  \\ \hline
		\end{tabular}
	\end{center}
	\vspace{-0.3cm}
	\caption{Effect of the $\delta$ value in the Huber loss function. Huber loss with $\delta=1$ is equivalent to smooth L1 loss.}	
	\label{tab:Loss}
\end{table}
\section{CONCLUSION}
\label{Conclusion}
The paper proposes a deep end-to-end network for disparity estimation in a multi-baseline trinocular setup. The pipeline processes two pairs of images from an axis-aligned three-camera configuration with narrow and wide baselines. In this regard, we introduced a new layer for effectively merging the information from the two baselines. A synthetic dataset was generated to help with a proposed iterative sequential learning of real and synthetic datasets. With this learning mechanism, we can train on a real dataset with no ground-truth information. Experiments show that the proposed method outperforms the disparity map estimated by each image pair. By providing improved flexibility and scalability, this multi-baseline deep model is promising for long range visual perception and autonomous navigation, where the image content is diversified and changeable in terms of the distance to the camera.
{\small
	\bibliographystyle{ieee_fullname}
	\bibliography{egbib}
}
\end{document}